\RequirePackage{fix-cm}
\documentclass[twocolumn,compsoc]{CVM}

\graphicspath{{figures/},{figures/cvm/}}
\DeclareGraphicsExtensions{.jpg,.pdf,.png,.eps}

\usepackage[pagebackref=false,breaklinks=false,letterpaper=false,colorlinks,bookmarks=false]{hyperref}
\usepackage{makecell}
\usepackage{overpic}
\usepackage{multirow}
\usepackage{bigstrut}
\usepackage{graphicx}
\usepackage{threeparttable}

\def\ie{\emph{i.e.}}
\def\eg{\emph{e.g.}}

\def\etal{\textit{et al.}}

\usepackage{pifont}% http://ctan.org/pkg/pifont
\usepackage{rotating}

\newcommand{\ourdataset}{\textit{ASOD60K}}
\newcommand{\ourtask}{PV-SOD}
\newcommand{\OurTotalFrames}{62,455}
\newcommand{\OurTotalObjects}{10,465}
\newcommand{\OurTotalInstances}{19,904}

\newcommand{\OurTotalVideos}{67}
\newcommand{\OurTotalBaselines}{11}

\newcommand{\yz}[1]{{\textcolor{black}{#1}}}

\def\ManuscriptInfo{Manuscript received: 2021-07-24; accepted: 2021-XX-XX.}

\def\PaperTitle{ASOD60K: Audio-Induced Salient Object Detection in Panoramic Videos}
\def\PaperTitle{\yz{ASOD60K: An Audio-Induced Salient Object Detection Dataset for 
Panoramic Videos}}

\def\AuthorNames{Yi Zhang$^{1}$}
\def\AuthorAdress{
    $1\quad $ & Institut National des Sciences Appliquées de Rennes (INSA Rennes), France. \\
}

\def\firstheader{DOI 10.1007/s41095-xxx-xxxx-x\hfill Vol. x, No. x, ** 2021, xx--xx\\[5mm]~Research Article}

\headevenname{{Y. Zhang}}

\begin{document}

\MakePageStyle

\MakeAbstract{
Exploring to what humans pay attention in dynamic panoramic scenes is useful for many fundamental applications, including augmented reality (AR) in retail, AR-powered recruitment, and visual language navigation. With this goal in mind, we propose \textbf{\ourtask}, a new task that aims to segment salient objects from panoramic videos. 
%In contrast to existing fixation-level or object-level saliency detection tasks, we focus on multi-modal salient object detection (SOD), which mimics human attention mechanism by segmenting salient objects with the guidance of audio-visual cues.
\yz{In contrast to existing fixation-/object-level saliency detection tasks, we focus on audio-induced salient object detection (SOD), where the salient objects are labeled with the guidance of audio-induced eye movements.}
To support this task, we collect the first large-scale dataset, named \ourdataset, which contains 4K-resolution video frames annotated with a six-level hierarchy, thus distinguishing itself with richness, diversity and quality. Specifically, each sequence is marked with both its super-/sub-class, with objects of each sub-class being further annotated with human eye fixations, bounding boxes, object-/instance-level masks, and associated attributes (\eg, geometrical distortion). These coarse-to-fine annotations enable detailed analysis for \ourtask~modeling, \eg, determining the major challenges for existing SOD models, and predicting scanpaths to study the long-term eye fixation behaviors of humans. We systematically benchmark \OurTotalBaselines~representative approaches on \ourdataset~and derive several interesting findings.
We hope this study could serve as a good starting point for advancing SOD research towards panoramic videos.
The dataset and benchmark will be made publicly available at \href{https://github.com/PanoAsh/ASOD60K}{https://github.com/PanoAsh/ASOD60K}.

}

\MakeKeywords{360° video salient object detection, saliency detection, panoramic videos, benchmark}

\begin{figure*}[t!]
	\centering
	\begin{overpic}[width=.98\textwidth]{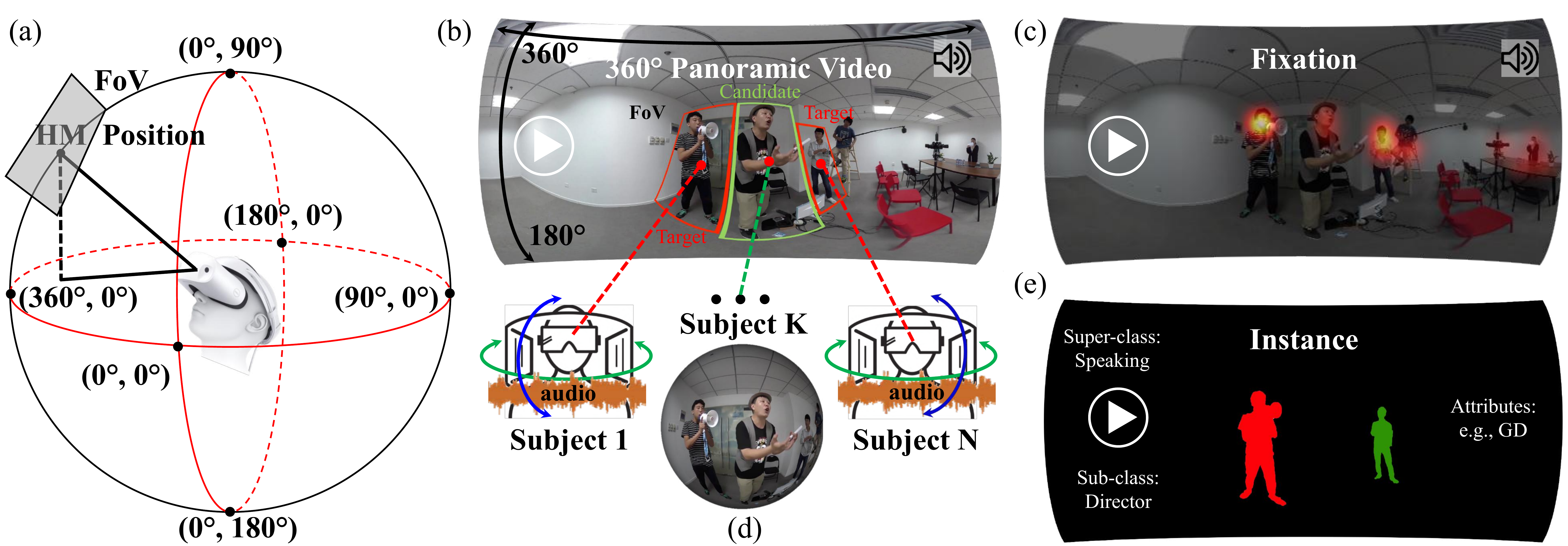}
    \end{overpic}
	\caption{\textbf{Annotation examples from the proposed \ourdataset~dataset}. 
    (a) Illustration of head movement (HM) \cite{pvshm}. The subjects wear Head-Mounted Displays (HMDs) and observe 360° scenes by moving their head to control a field-of-view (FoV) in the range of 360°$\times$180°.
    (b) Each subject (\ie, Subject 1 to Subject N) watches the video without restriction.
    (c) The HMD-embedded eye tracker records their eye fixations. 
    (d) According to the fixations, we provide coarse-to-fine annotations for each FoV including 
    (e) super/sub-classes, instance-level masks and attributes (\eg, GD-Geometrical Distortion).}
    \label{fig:AnnotationExamples}
\end{figure*}

\section{Introduction}\label{sec:intro}
%未来的通用人工智能需要机器人能够了解人的想法。例如自动驾驶的场景中，如果AI能够预判基于人眼对场景的感知，那么这样的机器人会显得更加人性化，让司机意思到潜在的危险，以及最可能关注到的物体。
%这是一个全新的开放问题，带着这样的研究动机，在这项工作中，我们花了一年的时间来构建这样的数据集、评估目前模型的性能，并且对未来的挑战和方向提出了建议。
Recently, AI companies and manufacturers have developed several panoramic cameras, such as Facebook's Surround360, Insta360 One, Ricoh Theta, and Google Jump VR, which produce omnidirectional\footnote{In the following sections, we use `omnidirectional', `panoramic', and `360°' interchangeably.} images (\figref{fig:AnnotationExamples} (b)) capturing a scene with a 360$^\circ\times180^\circ$ field-of-view (FoV). Thus, exploring human attention in dynamic scenes captured by these devices is of significant importance to augmented/virtual reality (AR/VR) applications, \eg, shopping, online recruitment, piloting~\cite{hu2017deep}, automatic cinematography~\cite{su2016pano2vid}, and immersive games. % (or spherical)

In practice, we have found that existing object-level saliency detection (\ie, SOD) techniques and the datasets that underpin their progress are subject to two 
important limitations. 
First, the input source only includes the \textit{visual information} from images (\eg, I-SOD (Image SOD)~\cite{fan2018salient,wang2021salient,borji2019salient,borji2015salient}, CoSOD~\cite{fan2020taking,Fan2021Group,deng2021re}, RGB-D SOD~\cite{zhou2021rgb,piao2019depth,liu2020learning,20TNNLS_RgbdBench,fu2021siamese,fu2020jl}, RGB-T SOD~\cite{tang2019rgbt,tu2019rgb}, LFSOD~\cite{jiang2020light,piao2020dut}, HRSOD~\cite{zeng2019towards,zhang2021looking}, Remote Sensing SOD~\cite{zhang2020dense}), or videos (\eg, V-SOD (Video SOD)~\cite{SSAV,ji2021fsnet,wang2021semantic}) ignoring the \textit{auditory cues} that are ubiquitous in dynamic scenes~\cite{tavakoli2019dave,Tsiami_2020_CVPR,van2008audiovisual}.
%
%Furthermore, the FoV in a 2D image/video is a plane rather than a sphere (\figref{fig:AnnotationExamples} (d)) as in a panoramic video, thus failing to capture the surrounding context and corresponding layout. This rich geometrical cue is crucial for object-level human attention modeling.
Furthermore, all the above mentioned SOD tasks focus on 2D images/videos, which are regarded as perspective images with local FoVs (compared with 360°$\times$180° FoV), thus failing to capture the surrounding context and corresponding layout of immersive real-life daily scenes.
However, these rich global geometrical cues are crucial for human attention modeling.

To this end, we envision that segmenting salient objects from panoramic videos with \textit{audio-visual} data will benefit not only our research community but also commercial products. 
To facilitate the study of \textbf{p}anoramic \textbf{v}ideo \textbf{s}alient \textbf{o}bject \textbf{d}etection (\textbf{\ourtask}), we collect \ourdataset\footnote{Collecting the six types of labels was a costly and time-consuming work, and it took us about 1 year to set up this large-scale database.}, the first large-scale \ourtask~dataset providing professional annotations. 
\ourdataset~has several 
%discriminative 
distinctive
features: 
\begin{itemize}
    \item \textit{Hierarchical categories.} All videos in the database are labeled in a hierarchical manner, \ie, with the \textit{super-class} and \textit{sub-class}. The two-level semantic categories provide a solid foundation for not only weakly supervised approaches but also fully supervised models. 
    
    \item \textit{Diverse annotations.} For each video sequence/frame, we 
    provide coarse-to-fine annotations, including the subjects' head movement (HM) and eye fixations, bounding boxes, object-level masks, and instance-level labels, which can greatly benefit different vision tasks (\eg, scanpath prediction, fixation prediction, SOD and salient instance detection).
    %, as shown in \figref{fig:AnnotationExamples}.
    
    \item \textit{Attribute labels.} Each sequence is annotated with specific attributes, \eg, \textit{geometrical distortion}, \textit{occlusions}, and \textit{motion blur}.
    Combined with the performance of the evaluated models, these attributes (\tabref{tab:Attributes} \& \figref{fig:AttExample})
    shed new light on the experimental analysis.
    
    \item \textit{High quality.} All video sequences are in high-resolution (4K) to adapt to VR devices such as Head-Mounted Displays (HMDs). Moreover, cross-checking (i.e., more than three-fold) by multiple experts and volunteers is conducted to maintain reliability, accuracy, and consistency during the whole annotation process.
\end{itemize}
The aforementioned aspects together provide important support for studying human attention in panoramic videos. Further, we summarize the design rules that a balanced \ourtask~dataset should fulfill, which can be used as reference for similar fields when collecting and labeling data.

%findings, HM highly concistent across different subjects?
%结论 
%1. 这个任务还有很大空间， 
%2. 受试之间的注意力在没有声源的情况下呈现很大偏差, 加入声音后，这些注视点高度一致
%3. 注意力转移是建模的难点
%4. 单纯的从图像的方法也能有很好的表现，说明视频中具有静态处理能力很重要，专门的视频设计还有很大的空间
To reveal the challenges of \ourtask, we perform a set of empirical studies 
based on the collected \ourdataset~dataset. We obtain three interesting observations. 
i) According to the overall ($S_\alpha<0.7$) and attribute-based performance of the tested models, this 
task is still far from being solved.
ii) We find that the eye fixations with audio are relatively consistent across subjects while the data without audio exhibit large fluctuations between different subjects.
iii) A sparsely labeled database is more beneficial for image-based models but more challenging for video-based models. 
%已经用image的train过的直接测试，以及已经用video，train过的，然后直接测试看看情况
%4) Models pre-trained on 2D image or video without audio for SOD cannot guarantee to achieve good generalizability. 
These findings clearly show the challenges of salient object detection in panoramic videos.

In a nutshell, our main contributions are twofold: 
i) We introduce \ourdataset, the first \textit{large-scale} dataset for PV-SOD, which consists of \OurTotalFrames~high-resolution (4K) video frames from \OurTotalVideos~carefully selected 360° panoramic video sequences.
\OurTotalObjects~key frames are annotated with rich labels, namely, \textit{super-class}, \textit{sub-class}, \textit{attributes}, \textit{HM data}, \textit{eye fixations}, \textit{bounding boxes}, \textit{object masks}, and \textit{instance masks}. 
ii) Based on the established \ourdataset, we present a comprehensive study on \OurTotalBaselines~representative models, which serves as the first standard leaderboard. Based on the evaluation results, we present insightful conclusions that may inspire novel ideas toward new research directions.

\begin{table*}[t!]
  \centering
  \renewcommand{\arraystretch}{1.0}
  \setlength\tabcolsep{2.5pt}
  \footnotesize
  \caption{
   Summary of widely used salient object detection (SOD) datasets and the proposed panoramic video SOD (PV-SOD) dataset. \#Img: The number of images/frames. \#GT: The number of ground-truth masks. Pub. = Publication. Obj.-Level = Object-Level. Ins.-Level = Instance-Level. Fix. GT = Fixation-guided ground truths. $\dag$ denotes equirectangular (ER) images.
   %Attr. = Attributes.
   %Note that all the datasets listed above provide pixel-wise annotations.
   }\label{tab:related works}
  %\resizebox{0.99\textwidth}{!}{
  \begin{tabular}{r|c|c|c|rr|rr|ccccc}
   \toprule
   Dataset~~~~~~~~ & Task & Year & Pub. & \#Img & \#GT  & $\texttt{min}(W,H)$ & $\texttt{max}(W,H)$ & Obj.-Level & Ins.-Level & Attribute & Fix. GT & Audio
  \\
  \hline
  \hline
  ECSSD~\cite{ECSSD} &  I-SOD & 2013 & CVPR 
  & 1,000 & 1,000 & 139 & 400 & \checkmark & & & \\
  DUT-OMRON \cite{DUTO} &  I-SOD & 2013 & CVPR 
  & 5,168 & 5,168 & 139 & 401 & \checkmark & & & \checkmark \\
  SegTrack V2 \cite{SegV2} &  V-SOD & 2013 & ICCV  
  & 1,065  & 1,065 & 212 & 640 & \checkmark & & & \\
  PASCAL-S \cite{PASCALS} &  I-SOD & 2014 & CVPR 
  & 850 & 850 & 139 & 500 & \checkmark & & & \checkmark \\
  FBMS \cite{FBMS} &  V-SOD & 2014 & TPAMI 
  & 13,860 & 720 & 253 & 960 & \checkmark & & & \\
  HKU-IS \cite{HKUIS} &  I-SOD & 2015 & CVPR 
  & 4,447 & 4,447 & 100 & 500 & \checkmark & & &  \\
  MCL~\cite{kim2015spatiotemporal} & V-SOD & 2015 & TIP 
  & 3,689 & 463 & 270 & 480 & \checkmark & & & \\
  ViSal \cite{ViSal} &  V-SOD & 2015 & TIP 
  & 963 & 193 & 240 & 512 & \checkmark & & & \\
  DAVIS2016 \cite{DAVIS} &  V-SOD & 2016 & CVPR 
  & 3,455 & 3,455 & 900 & 1,920 & \checkmark & & \checkmark & \\
  DUTS \cite{DUTS} &  I-SOD & 2017 & CVPR
  & 15,572 & 15,572 & 100 & 500 & \checkmark & & &  \\
  ILSO \cite{li2017instance} &  I-SOD & 2017 & CVPR 
  & 1,000 & 1,000 & 142 & 400 & \checkmark & \checkmark & &  \\
  UVSD~\cite{liu2017saliency} & V-SOD & 2017 & TCSVT
  & 3262 & 3262 & 240 & 877 & \checkmark & & & \\
  SOC \cite{fan2018salient} &  I-SOD & 2018 & ECCV
  & 6,000 & 6,000 & 161 & 849 & \checkmark & \checkmark & \checkmark & \\
  VOS \cite{VOS} &  V-SOD & 2018 & TIP 
  & 116,103 & 7,467 & 312 & 800 & \checkmark & & & \checkmark\\
  DAVSOD \cite{SSAV} &  V-SOD & 2019 & CVPR 
  & 23,938 & 23,938 & 360 & 640 & \checkmark & \checkmark & \checkmark & \checkmark \\
  %\hline
  F-360I-SOD \cite{Yi2020fSOD}  & PI-SOD & 2020 & ICIP 
  & 107$^\dag$ & 107 & 1,024 & 2,048 & \checkmark & \checkmark &  & \checkmark
  \\
  360-SOD \cite{li2020distortion} & PI-SOD & 2020 & JSTSP 
  & 500$^\dag$ & 500 & 512 & 1,024 & \checkmark & & & \\
  360-SSOD \cite{ma2020stage} & PI-SOD & 2020 & TVCG 
  & 1,105$^\dag$ & 1,105 & 546 & 1,024 & \checkmark & & & \\
%   \\
  \hline
  \textbf{\ourdataset~(OUR)} & \textbf{PV-SOD} & 2021 & CVMJ 
  & \OurTotalFrames$^\dag$ & \OurTotalObjects & \textbf{1,920} & \textbf{3,840} & \checkmark & \checkmark & \checkmark & \checkmark & \checkmark \\
%   \\
  \bottomrule
  \end{tabular}
  %}
\end{table*}

\section{Related Work}\label{sec:related_work}
Human attention modeling in panoramic videos can be roughly split into four levels: HM prediction~\cite{pvshm}, eye fixation/gaze prediction~\cite{wu2020spherical,xu2018gaze}, salient object detection (SOD), and salient instance detection. Our work mainly focuses on the object-level task, leaving other tasks to our future studies. 
In this section, we only briefly discuss some closely related works, \ie, datasets, models, and techniques for 360° image processing.

\subsection{Datasets}
\noindent
\textbf{Image Salient Object Detection (I-SOD).} The image-based SOD task has gained significant attention in the past few years. The remarkable progress of I-SOD is highly related to the development of several representative datasets~\cite{ECSSD,DUTO,PASCALS,HKUIS,DUTS,li2017instance,fan2018salient}. ECSSD~\cite{ECSSD}, DUT-OMRON~\cite{DUTO}, PASCAL-S~\cite{PASCALS}, and HKU-IS~\cite{HKUIS} are four early, small-scale datasets with limited image resolution. To increase the amount of training data, DUTS~\cite{DUTS} was introduced and has become one of the most popular benchmarks. Furthermore, ILSO~\cite{li2017instance} and SOC~\cite{fan2018salient} were recently proposed with the goal of enabling not only object-level but also instance-level SOD tasks.
We refer the reader to the survey paper by~\cite{wang2021salient} for a thorough review. 

\noindent
\textbf{Video Salient Object Detection (V-SOD).} 
In addition to I-SOD datasets, several V-SOD benchmarks have also been introduced. \tabref{tab:related works} summarizes their details. As can be seen, DAVSOD~\cite{SSAV} is the largest dataset and provides comprehensive annotations for the V-SOD task.

\noindent
\textbf{Panoramic Image Salient Object Detection (PI-SOD).} 
There are three attempts toward establishing datasets for 360° panoramic image-based SOD~\cite{li2020distortion,ma2020stage,Yi2020fSOD}, all of which provide pixel-wise object-level ground truths (GTs) with similar resolution (\eg, $\texttt{max}(w,h)=2,048$). 
360-SOD~\cite{li2020distortion} is the pioneering work for SOD in 360$^\circ$ scenes. It consists of 500 equirectangular (ER) images (the most widely used planar representation of 360° image without any loss of spatial information) representing both the indoor/outdoor scenes with object-level annotations. 360-SSOD~\cite{ma2020stage} is 
a larger public PI-SOD dataset that has 1,105 semantically balanced panoramic (ER) images. Besides, F-360iSOD \cite{Yi2020fSOD} is so far the only 360° image SOD dataset that provides pixel-wise instance-level GTs. 

%360° image: Salient!360
%360° video: PVS-HMEM, 360VHMD, Wild-360, VR-Scene, 360-Saliency
\noindent
\textbf{Other Datasets.} 
Other closely related datasets are either designed to simulate human HM (360VHMD~\cite{corbillon2017360} and PVS-HMEM~\cite{pvshm}), or eye fixations (\ie, saliency detection) in 360°/panoramic images (Salient!360~\cite{rai2017dataset}) or 360° videos (\eg, Wild-360~\cite{cheng2018cube}, VR-Scene~\cite{xu2018gaze}, 360-Saliency~\cite{zhang2018saliency}). As far as we know, Chao \etal's\cite{chao2020audio} work is the only 360$^\circ$ audio-visual saliency dataset, which contains 12 videos with HM-based annotations under mute, mono, and ambisonics modalities. Finally, datasets such as \cite{yang2018object,360indoorWACV2020,360sports} focus on bounding-box-level object detection in 360°.
%
%However, no efforts made in the community for panoramic video salient object detection (PV-SOD).

As summarized in \tabref{tab:related works}, no work exists to study the segmentation of salient objects in free-viewing panoramic videos with audio. The closest works are audiovisual saliency detection from 2D videos~\cite{Tsiami_2020_CVPR,jain2020avinet} in a plane and salient object detection in omnidirectional images~\cite{li2020distortion,Yi2020fSOD,ma2020stage} \textit{without audio}. We refer readers to the survey paper about 360° data processing~\cite{xu2020state,fan2019survey} for more details.

\subsection{SOD Models}\label{sec:SOD_method}

Since no SOD approaches currently exist for the PV-SOD task, we present the SOD methodologies for I-SOD, V-SOD, and PI-SOD.

\noindent
\textbf{Algorithms for I-SOD}. In the past few years, convolutional neural networks (CNNs) have been the most commonly used architecture in state-of-the-art (SOTA) I-SOD models~\cite{ASNet,AFNet,BASNet,qin2021boundary,zhuge2021salient,CPD,PoolNet,EGNet,SCRN,F3Net,GCPANet,ITSD,MINet,SOD100K}, which are trained on large-scale datasets (\eg, DUTS~\cite{DUTS}) in a fully supervised manner. ASNet~\cite{ASNet} uses eye fixations to aid salient object localization, while models such as AFNet \cite{AFNet}, BASNet \cite{qin2021boundary}, PoolNet \cite{PoolNet}, EGNet \cite{EGNet}, SCRN \cite{SCRN}, and LDF\cite{CVPR2020LDF} emphasize object appearance (boundaries or skeletons) as guidance for the accurate segmentation of salient objects. With comparable accuracy, methods such as CPD \cite{CPD}, ITSD \cite{ITSD}, and CSNet \cite{SOD100K} also achieve improved inference speed. %With the most recent development of image datasets such as CoSOD3k \cite{fan2020taking} and CoCA \cite{GICD}, methods specially designed (\eg, GICD \cite{GICD}) for the co-saliency detection have also been proposed.

\noindent
\textbf{Models for V-SOD}. The recent development of large-scale video datasets such as DAVIS \cite{DAVIS} and DAVSOD \cite{SSAV} have enabled deep learning-based V-SOD. Several works~\cite{MGA,li2018flow,RCRNet} have achieved success by introducing optical flow cues into the network. There is, however, the long-standing and often ignored issue of \textit{saliency shift}, which was first highlighted and modeled 
in SSAV~\cite{SSAV}. According to the open benchmark results, COSNet~\cite{COSNet}, RCRNet~\cite{RCRNet}, and PCSA~\cite{gu2020PCSA} obtain best performances in the V-SOD task.

%\cite{MGA}, \cite{li2018flow} and \cite{RCRNet} modeled the temporal information by combining optical flow. 
%SSAV \cite{SSAV} discovered and applied a fixation-based attention shift mechanism to help recognizing moving salient objects. 
%COSNet \cite{COSNet}, AGS \cite{AGS} and RANet \cite{RANet} developed weight-sharing modules trained with video frames from multiple input channels.

\begin{figure*}[t!]
	\centering
	\begin{overpic}[width=.98\textwidth]{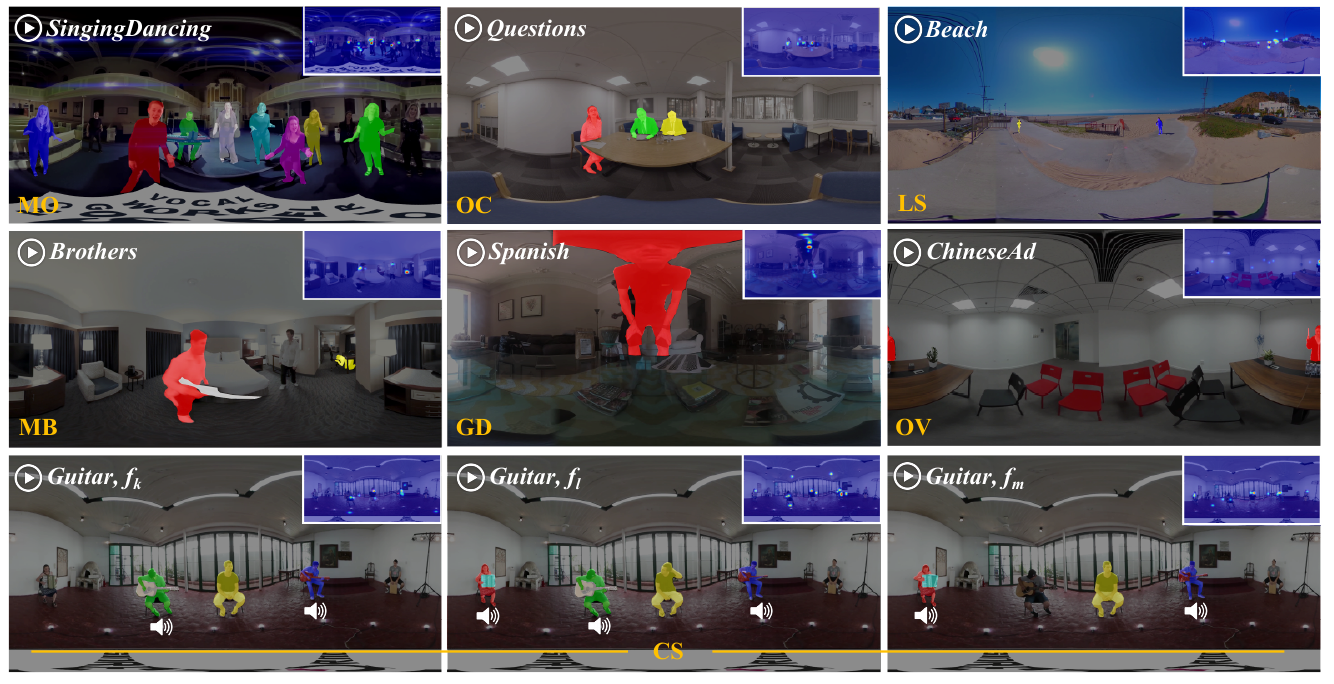}
    \end{overpic}
	\caption{
	Examples of challenging attributes (see \tabref{tab:Attributes}) on ER images from our \ourdataset, with instance-level GT and fixations as annotation guidance. $f_{k,l,m}$ denote random frames of a given video. Best viewed in color. More examples are \yz{shown} in \figref{fig:show_part1}.}
    \label{fig:AttExample}
\end{figure*}

\noindent
\textbf{Methods for PI-SOD}. To the best of our knowledge, DDS \cite{li2020distortion}, stage-wise SOD \cite{ma2020stage} and FANet \cite{huang2020fanet} are so far the only models exclusively designed for PI-SOD. They all emphasize the importance of mitigating the geometrical distortion brought by ER projection via specific modules.

\subsection{360° Image Processing Approaches}

The vast majority of current 360° image processing techniques are CNN-based, proposed for either ER or stereoscopic images (\eg, spheres and icosahedrons).

\noindent
\textbf{CNNs on ER Images.} ER projection is the most widely used approach for the 2D representation of a 360$^\circ$ image. It applies a uniform grid-based sampling method on the spherical surface, followed by an inevitable over-sampling of spherical regions near poles. 
Therefore, salient objects in ER images may suffer geometrical distortions to varying extents, depending on the distance between their geometrical locations and the equator of the ER image (\figref{fig:AttExample}). SphereNet \cite{coors2018spherenet}, which consists of a location-adaptive kernel, was proposed for the classification and detection of objects in ER images. Similar location-dependent convolutional kernels are also applied in~\cite{su2017learning,su2019kernel}.

\noindent
\textbf{CNNs on Stereoscopic Images}. As there is no perfect 2D representation for 360$^\circ$ images, SO(3)-based spherical CNNs~\cite{cohen2018spherical,esteves2018learning} were proposed to directly generalize convolutions on a sphere. However, these reparameterized 3D convolutional kernels hinder the use of classical backbones (\eg, ResNet~\cite{he2016deep} or VGGNet~\cite{Simonyan15}) pre-trained on large-scale datasets (\eg, ImageNet~\cite{deng2009imagenet}), which play an essential role in CNN-based SOD models (see \secref{sec:SOD_method}). Recent researches~\cite{jiang2019spherical,cohen2019gauge,zhang2019orientation} generalize convolutions on subdivided icosahedral faces, which contain much less geometrical distortion compared to ER images~\cite{eder2020tangent}. 
In addition, tangent image \cite{eder2020tangent} was proposed to enable the implementation of semantic segmentation in 4K-resolution 360$^\circ$ images.

\section{Proposed ASOD60K Dataset}\label{sec:AV360}

We elaborate our \ourdataset~in terms of stimuli collection, subjective experimentation, annotation pipeline and dataset statistics. Our goal is to introduce a new challenging dataset to the PV-SOD community.

%\subsection{Key Challenges}
%目标数据集都是fixation的，要标注物体比较难，我们的策略是什么
%收集有声音的不容易？
%如何定位一个物体和下一个物体，看ssav的反馈意见
%如何标注属性

\subsection{Stimuli Collection}

The stimuli of \ourdataset~were searched on \textit{YouTube} with different keywords (\eg, 360°/panoramic/omnidirectional video, spatial audio, ambisonics \cite{morgadoNIPS18}). As a result, our collected stimuli cover various real-world scenes (\eg, indoor/outdoor scenes), different occasions (\eg, sports, travel, concerts, interviews, dramas), different motion patterns (\eg, static/moving camera), and diverse object categories (\eg, humans, instruments, animals). They possess a wide range of major challenges found in 360° content\footnote{Objects scattered far from the equator thus suffering from serious geometrical distortions in ER projections.}, providing us with a solid foundation to build a representative benchmark.
In this way, we obtained about 1,000 \yz{noisy} videos, \eg, videos with a shaking camera, dark-screen transitions, without key content, displaying too many objects, of low quality. 
In line with the video dataset creation rules in~\cite{SSAV,wang2018revisiting}, we then carefully collected \OurTotalVideos~high-quality video sequences with a total of \OurTotalFrames~frames recorded with \OurTotalFrames$\times$40 HM and eye fixations.
\yz{The \OurTotalVideos~sequences are selected based on the following two criteria, 1) The source video must be in good visual quality, \ie, 4K resolution for each video frame. 2) The scenes must include meaningful objects on which high saliency constantly focuses. In other words, scenes such as busy streets and carnivals where subjects' attention simultaneously scattering on multiple non-relevant objects, are removed from the dataset.}
Similar to~\cite{corbillon2017360}, the frame rate of each collected video is not fixed (varying from 24fps to 60fps), which did not influence the results of following subjective experiments since human attention is mainly event-related, rather than frame rate-dependent. Note that we manually trimmed the videos into small clips (29.6s on average) to avoid fatigue during the collection of human eye fixations. As a result, the final duration is 1983s in total.

%but this does not impact the data collection since the eye-fixation recordings are based on each event in the video rather than a fixed rate.
%, with the eye-fixation records based on particular events in video rather than at a fixed rate.
%

\subsection{Subjective Experimentation}\label{sec:subjectExp}

\noindent
\textbf{Equipment.}
All the video clips were displayed using a HTC Vive HMD embedded with a Tobii eye tracker with 120Hz sample rate to collect eye fixations. %in our subjective experiments. 

\noindent
\textbf{Observers.}
We recruited 40 participants (8 females and 32 males) aging from 18 to 34 years old who reported normal or corrected-to-normal visual and audio acuity.
Twenty participants were randomly selected to watch videos with mono sound, while the other participants watched videos without sound. %
Note that the two groups own the same gender and age distributions. 
Hence, each video with each audio modality (\ie, with or without sound) was viewed by 20 participants, and each participant viewed each video only once. We performed task-free viewing sessions. %in our experiments. 

\noindent
\textbf{Settings.}
All the participants seated in a swivel chair, wearing a HMD with headphones, and asked to explore the panoramic videos without any specific intention. 
During the experiments, the starting position was fixed to the center ($\theta=0^{\circ}$ and $\phi=0^{\circ}$) at the beginning of every video display. To avoid motion sickness and eye fatigue, we inserted a short rest of a five-second gray screen between two successive videos and a long break of 20 minutes after every 20 videos. 
We calibrated the system for each participant at the beginning and end stage of every long break.

\subsection{Professional Annotation}

\begin{figure*}[t!]
	\centering
	\begin{overpic}[width=1\textwidth]{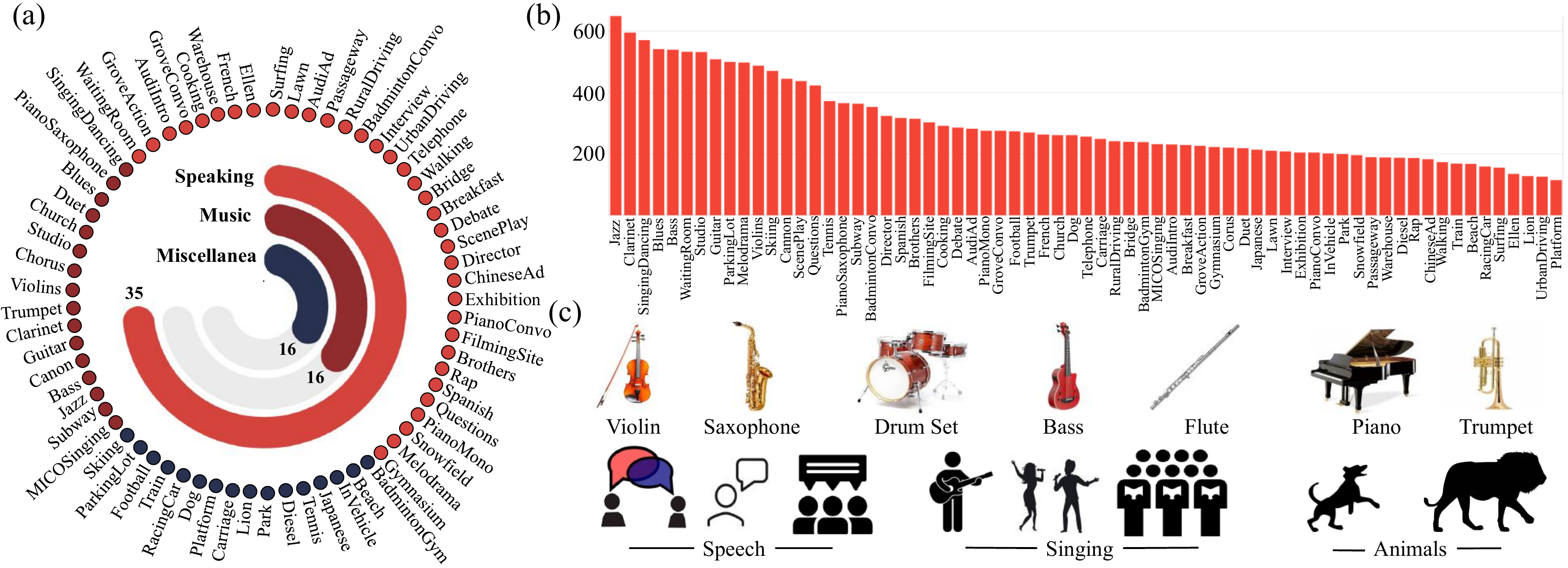}
    \end{overpic}
	\caption{
	Statistics of the proposed \ourdataset. (a) Super-/sub-category information. (b) Instance density of each sub-class. (c) \yz{Main sound sources of \ourdataset~scenes, such as musical instruments, human instances and animals.} Best viewed in color.}
    \label{fig:categories}
\end{figure*}

\noindent
\textbf{Super-/Sub-Class Labeling.}
As shown in \figref{fig:categories}, our \ourdataset~contains 67 videos representing three super-categories of audio-introduced scenes, 
including speaking (\eg, monologue, conversation), music (\eg, human singing, instrument playing) 
and miscellanea (\eg, the sound of vehicle engines and horns on the streets, crowd noise in the open air). 
Each video is named in terms of its audio-visual information.

\noindent
\textbf{Head Movement and Eye Fixations.} 
The recent video object segmentation dataset DAVIS \cite{DAVIS} contains only one or several foreground objects per frame, where the salient objects can be easily defined. In contrast, other recent video SOD datasets, such as VOS \cite{VOS} and DAVSOD \cite{SSAV}, collect video stimuli representing more challenging scenes with multiple salient objects. In such cases, fixation-based annotations (\eg, saliency-shift \cite{SSAV}) are used as guidance to alleviate the ambiguity of defining salient objects. 
Based on the subjects' per-frame HM and eye fixations gained by conducting the subjective experiments (\secref{sec:subjectExp}) with audio-visual stimuli, we produced the final annotations. \yz{Specifically, inspired by the experimental IoU threshold, \ie, AP50 (threshold set as 50\%), which is widely used in the field of object detection, we choose to keep the 50\% of the smoothed fixation map regions which thus covers the top 50\% of the saliency. In this way, we regard the top 50\% saliency as high saliency and locate the salient objects overlapped with the high saliency regions.}

\begin{table}[b!]
   \footnotesize
   \renewcommand{\arraystretch}{1.0}
   \setlength\tabcolsep{5pt}
   \caption{
  Attributes description (see examples in  \figref{fig:AttExample}).}\label{tab:Attributes}
 %\resizebox{0.46\textwidth}{!}{
 \begin{tabular}{ll}
  \hline
  \toprule
  Att. & Description \\
  \hline
  \textbf{MO} & \emph{Multiple Objects.} $\geqslant$ three objects occur simultaneously.\\
  \textbf{OC} & \emph{Occlusions.} Object is partially occluded.\\
  \textbf{LS} & \emph{Low Space.} Object occupies \textbf{$\leqslant0.5\%$} of image area.\\
  \textbf{MB} & \emph{Motion Blur.} Moving object with fuzzy boundaries.\\
  \textbf{OV} & \emph{Out-of-View.} Object is cut in half in ER projection.\\
  \textbf{GD} & \emph{Geometrical Distortion.} Distorted object in ER projection.\\
  \textbf{CS} & \emph{Competing Sounds.} Sound objects compete for attention.\\
  \hline
  \toprule
  \end{tabular}
 % }
\end{table}

 %动机为什么要用boundingbox 参考一下我那篇论文
 %做法1. 投影
% To obtain reliable bounding box annotations, we first project the xxx to xxx image and overlap xxx.
 %
 %Then, we identify the salient objects with eye-fixation-based pixel-wise $(x,y)$ high saliency ($S_{x,y} \geqslant 0.5$, where $S_{x,y} \subseteq [0, 1]$). 
%2.软件标注
%3.说明结果，最后标注了多少张
%xxxx
%--------------------------------bounding box 标注流程---------------------------------
%介绍360°物体检测常用的两种标注；说明选择bbox的动机
\noindent
\textbf{Bounding Box Annotations.}
Generally, there are two types of labels in 360° object detection, \ie, bounding FoVs~\cite{zhao2020spherical, 360indoorWACV2020, yang2018object} and bounding boxes~\cite{yang2018object}. As the vast majority of our collected video frames contain multiple salient objects near the 360 camera, bounding FoVs may introduce serious annotation ambiguities due to the divergence of projection angle selections between multiple annotators \cite{yang2018object}, thus not being suitable for the salient object annotation in the scenes from our ASOD60K. Following \cite{yang2018object}, we directly annotated the salient objects with bounding boxes in ER images.

%详细说明如何进行bbox标注
Our annotation protocol is threefold: 
i) We uniformly extracted \OurTotalObjects~key frames from the total \OurTotalFrames~frames with a sampling rate of 1/6. 
ii) We filtered the Gaussian-smoothed eye fixation maps corresponding to each of the key frames.
%, and thereshold them to keep only the top 50$\%$ high saliency (inspired by the common rule in the field of object detection: IoU$\geq$0.5).
iii) We adopted the widely used CVAT toolbox 
%\url{https://github.com/openvinotoolkit/cvat} 
as our annotation platform, and recruited an expert to manually annotate the bounding box of each salient object in each of the ER key frames, under the guidance of the corresponding fixations overlaid (see \figref{fig:AttExample}).
%标注结果
Finally, we obtained total \OurTotalInstances~salient objects labeled with instance-level bounding boxes from \OurTotalObjects~key frames. To the best of our knowledge, this is the first attempt to annotate salient objects with the guidance of audio-visual attention data.

%---------------------------------------------------------------------
\noindent
\textbf{Object-Level Annotations.}
%We uniformly sample \OurTotalObjects~key frames from the total \OurTotalFrames~video frames with a sampling rate of one out of six. 
With the coarse annotations (\ie, bounding boxes) in hand, we needed to further label the data in a fine manner.  
Thus, three experts were recruited to manually annotate the salient objects in the $\sim$10K key frames. To ensure satisfying annotations, they were first required to pass a training session\footnote{Note that it took the experts about 10 hours.} during which they had to correctly segment (by finely tracing objects' boundaries rather than drawing rough polygons) all the salient objects in a given video (previously shown to three senior researchers, with GTs acquired by consistent opinions), with the guidance of overlaid per-frame bounding boxes. Followed by a session during which they were asked to annotate all the defined salient objects in the rest of the panoramic videos.
Finally, a thorough inspection was conducted by the same three senior researchers, to ensure the accuracy of the annotations. 
Following the same pipeline as \cite{deng2021re}, we obtained \OurTotalObjects~object annotations.

\begin{figure*}[t!]
	\centering
	\begin{overpic}[width=\textwidth]{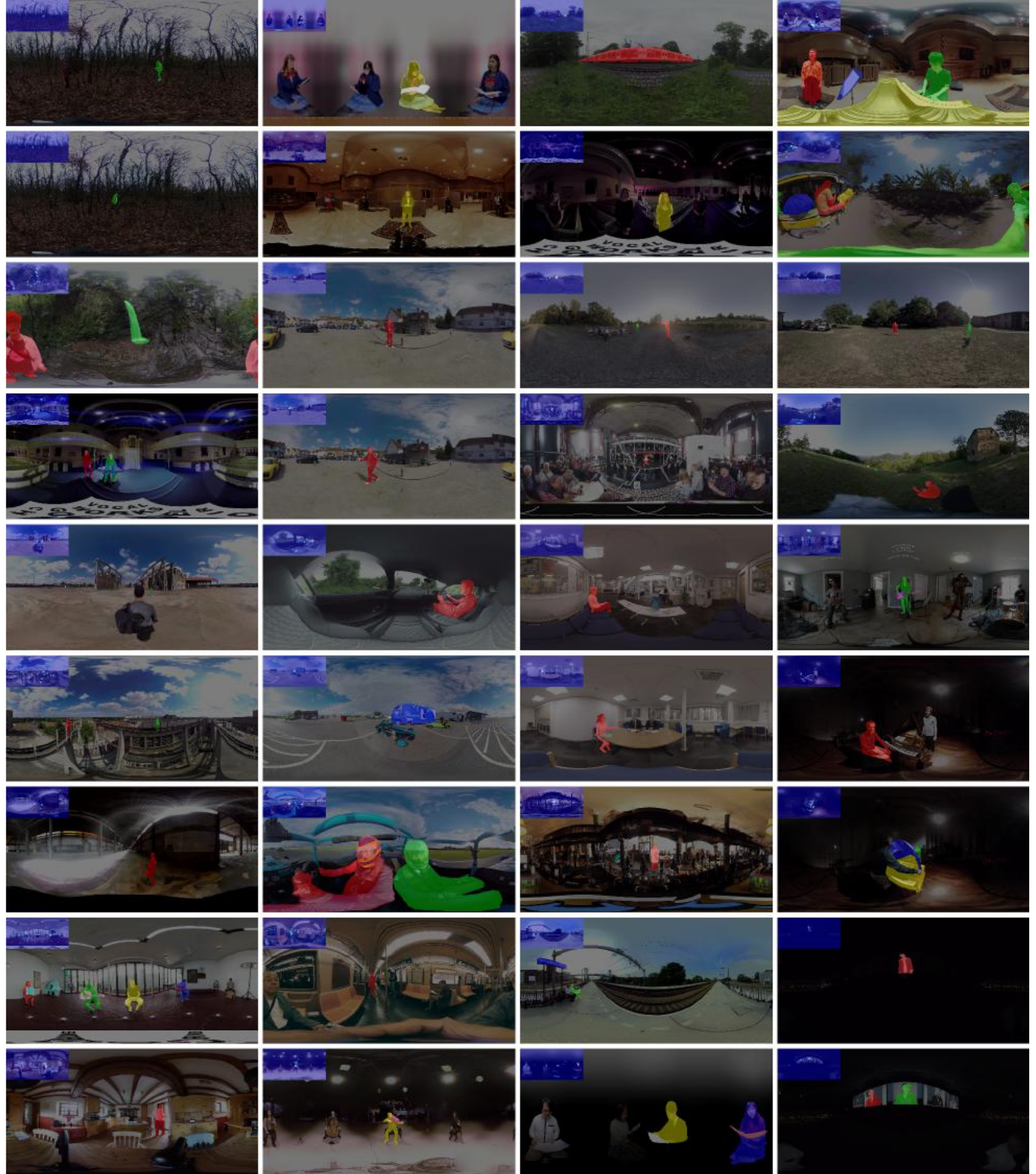}
    \end{overpic}
	\caption{Sample key frames from \ourdataset, with fixations and instance-level ground truth overlaid.}\label{fig:show_part1}
\end{figure*}

\noindent
\textbf{Instance-Level Annotations.}
Another three well-trained experts were then recruited to further draw pixel-wise instance-level masks by carefully tracing boundaries (rather than rough polygons) of the defined salient objects in each of the \OurTotalObjects~key frames. To ensure high quality annotations, all the masks were sent to a quality check procedure implemented by the same three senior researches. 
As a result, we acquired \OurTotalInstances~instance-level masks representing all the salient objects in all the \OurTotalObjects~key frames (the number of instances in each video are shown in \figref{fig:categories} (b)). 
Further, to refine the annotations quality, we transferred all the instance-level masks to object-level binary masks. 
The bounding boxes were also refined by the object-level masks.
%implement image/video SOD, we transfer all the instance-level masks to object-level binary masks. 
%Therefore, we finally acquire \OurTotalObjects~360° images with the corresponding \OurTotalObjects~object-level pixel-wise labels, which are regarded as the GT of our following benchmark studies.
Please refer to \figref{fig:show_part1} for annotation examples.

\begin{figure}[t!]
	\centering
    \begin{overpic}[width=\columnwidth]{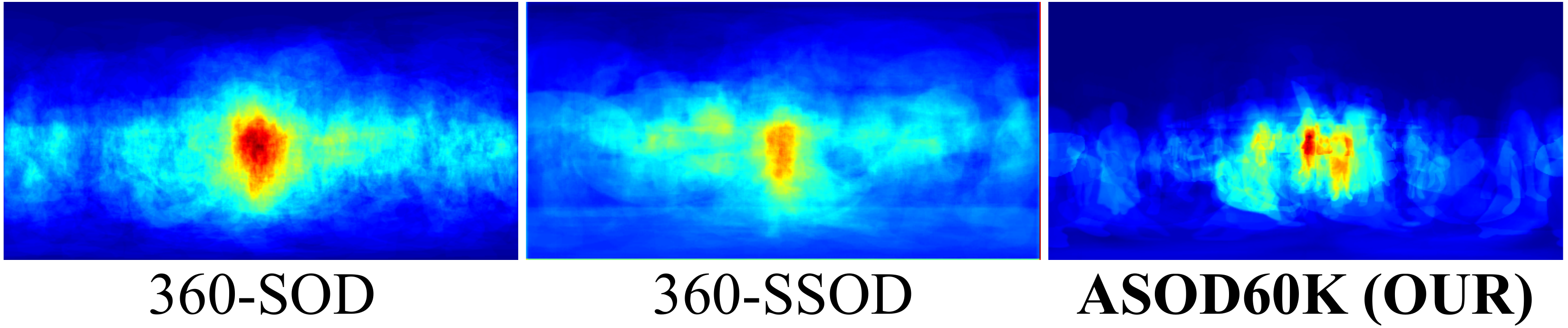}
    \end{overpic}
    \caption{
    Average object-level GT maps of 360-SOD \cite{li2020distortion}, 360-SSOD \cite{ma2020stage} and our \ourdataset.}\label{fig:GTComparison}
\end{figure}

\begin{figure}[t!] 
	\centering
	\begin{overpic}[width=\columnwidth]{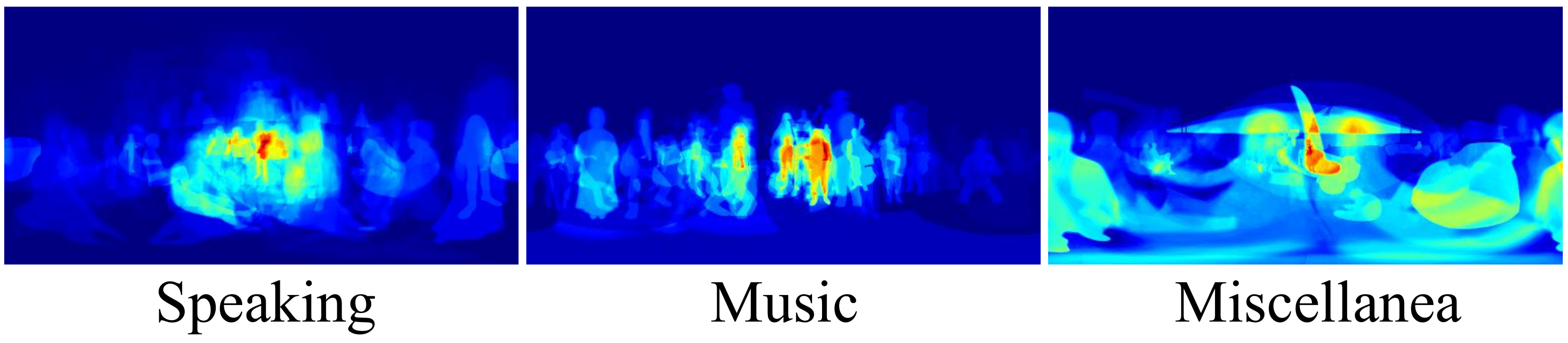}
    \end{overpic}
	\caption{The average object-level GT maps of our \ourdataset~at super-class level.}\label{fig:center_bias_class}
\end{figure}

\begin{figure}[b!]
	\centering
    \begin{overpic}[width=.98\columnwidth]{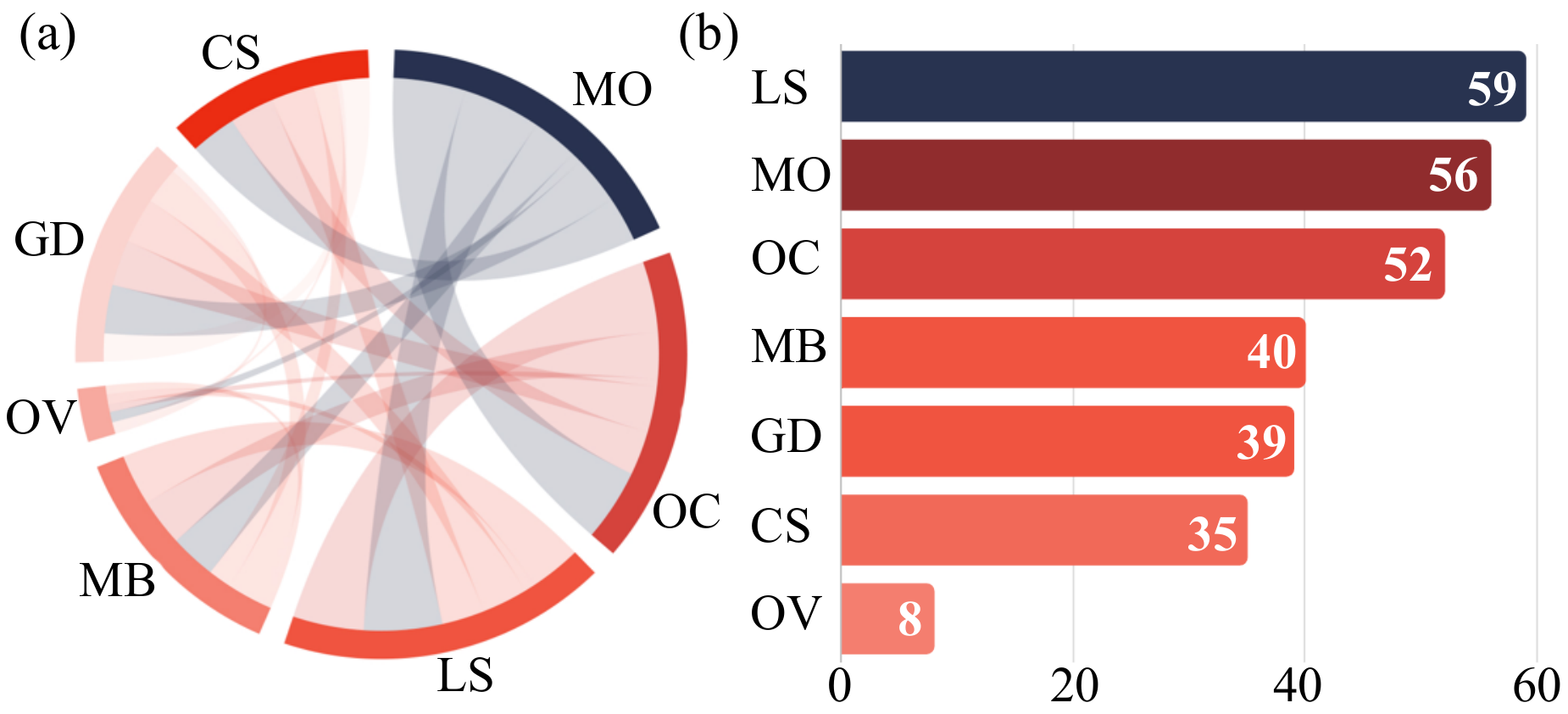}
    \end{overpic}
    \caption{
    Attributes statistics. (a)/(b) represent the correlation and frequency of \ourdataset's attributes, respectively.}
    \label{fig:attributes_corr}
\end{figure}
\noindent
\textbf{Attribute Labels.}
Following two large-scale video object segmentation datasets~\cite{DAVIS,SSAV}, we also provide seven attributes in the proposed \ourdataset, including \emph{multiple objects (MO)}, \emph{occlusions (OC)}, \emph{low space (LS)}, \emph{out-of-view (OV)}, \emph{motion blur (MB)}, \emph{geometrical distortion (GD)} and \emph{competing sounds (CS)} (Table \ref{tab:Attributes}). It is worth mentioning that, \emph{OV} and \emph{GD} (\figref{fig:AttExample}) are exclusive geometrical attributes of ER images, and \emph{CS} is a novel attribute attached to audio-visual stimuli (Please see per-video attributes' statistics in \tabref{tab:attributes_details}). 

\begin{table}[t!]
   \small
  \centering
  \renewcommand{\arraystretch}{0.3}
  \setlength\tabcolsep{2pt}
   \caption{Attribute details. General attributes: MO = multiple objects. OC = occlusions. LS = low space. MB = motion blur. 360° geometrical attributes: OV = out-of-view. GD = geometrical distortion. Spatial audio attributes: CS = competing sounds.}
   \label{tab:attributes_details}
  %\resizebox{0.47\textwidth}{!}{
  \begin{tabular}{cl||cccc|cc|c|c}
   \toprule
   \multicolumn{2}{l||}{\multirow{2}{*}{{Sequence}}}& \multicolumn{4}{c|}{General} & \multicolumn{2}{c|}{360°} & Audio & \multirow{2}{*}{{No.}}\\
   \cline{3-9}\\
   & & MO & OC & LS & MB & OV & GD & CS 
   \\
 \midrule
  \multirow{35}{*}{\begin{sideways}Speaking (35)\end{sideways}} &
  French & \ding{52} & \ding{52} & \ding{52} & \ding{52} & & \ding{52} & & 5 \\
  & WaitingRoom & \ding{52} & \ding{52} & \ding{52} & \ding{52} & & & \ding{52} & 5 \\
  & Cooking & \ding{52} & \ding{52} & \ding{52} & \ding{52} & & & \ding{52} & 5 \\
  & AudiIntro & \ding{52} & & \ding{52} & & \ding{52} &  & & 3\\
  & Ellen & & & \ding{52} & & & & & 1 \\
  & GroveAction & \ding{52} & \ding{52} & \ding{52} & \ding{52} & & & \ding{52} & 5 \\
  & Warehouse & & & \ding{52} & \ding{52} & & & & 2 \\
  & GroveConvo & \ding{52} & \ding{52} & \ding{52} & \ding{52} & &  & \ding{52} & 5 \\
  & Surfing & & \ding{52} & & \ding{52} & & \ding{52} & & 3 \\
  & Passageway & \ding{52} & & \ding{52} & \ding{52} & \ding{52} & & & 4\\
  & RuralDriving & \ding{52} & \ding{52} & \ding{52} & & & \ding{52} & & 4 \\
  & Lawn & & & & \ding{52} & & \ding{52} & & 2 \\
  & AudiAd & \ding{52} & \ding{52} & \ding{52} & \ding{52} & \ding{52} & \ding{52} & & 6 \\
  & ScenePlay & \ding{52} & \ding{52} & \ding{52} & & & \ding{52} & \ding{52} & 5 \\
  & UrbanDriving & \ding{52} & & \ding{52} & & & \ding{52} & & 3 \\
  & Interview & \ding{52} & \ding{52} & \ding{52} & & & & \ding{52} & 4 \\
  & Telephone & \ding{52} & \ding{52} & \ding{52} & \ding{52} & & \ding{52} & & 5 \\
  & Walking & & & \ding{52} & \ding{52} & & \ding{52} & & 3 \\
  & Bridge & & \ding{52} & \ding{52} & \ding{52} & & & \ding{52} & 4 \\
  & Breakfast & \ding{52} & \ding{52} & \ding{52} & \ding{52} & & \ding{52} & & 5 \\
  & Debate & \ding{52} & & \ding{52} & & & & \ding{52} & 3 \\
  & BadmintonConvo & \ding{52} & \ding{52} & \ding{52} & \ding{52} & \ding{52} & \ding{52} & \ding{52} & 7 \\
  & Director & \ding{52} & \ding{52} & \ding{52} & \ding{52} & & \ding{52} & \ding{52} & 6 \\
  & ChineseAd & \ding{52} & \ding{52} & \ding{52} & \ding{52} & \ding{52} & \ding{52} & & 6 \\
  & Exhibition & & & \ding{52} & & & & & 1 \\
  & PianoConvo & \ding{52} & & & & & \ding{52} & \ding{52} & 3 \\
  & FilmingSite & \ding{52} & \ding{52} & \ding{52} & \ding{52} & & & \ding{52} & 5 \\
  & Brothers & \ding{52} & \ding{52} & \ding{52} & \ding{52} & & \ding{52} & \ding{52} & 6\\
  & Rap & \ding{52} & \ding{52} & \ding{52} & \ding{52} & & & & 4 \\
  & Spanish & \ding{52} & \ding{52} & \ding{52} & \ding{52} & & \ding{52} & & 5 \\
  & Questions & \ding{52} & \ding{52} & \ding{52} & & & & \ding{52} & 4 \\
  & PianoMono & \ding{52} & \ding{52} & \ding{52} & \ding{52} & & \ding{52} & & 5 \\
  & Snowfield & & & & \ding{52} & & \ding{52} & \ding{52} & 3 \\
  & Melodrama & \ding{52} & \ding{52} & \ding{52} & & & \ding{52} &\ding{52} & 5 \\
  & Gymnasium & \ding{52} & \ding{52} & \ding{52} & \ding{52} & & \ding{52} & & 5 \\
  \midrule
  \multirow{16}{*}{\begin{sideways}Music (16)\end{sideways}} &
  Guitar & \ding{52} & \ding{52} & \ding{52} & & & & \ding{52} & 4 \\
  & Subway & \ding{52} & \ding{52} & \ding{52} & \ding{52} & & \ding{52} & & 5 \\
  & Jazz & \ding{52} & \ding{52} & \ding{52} & & & \ding{52} & \ding{52} & 5 \\
  & Bass & \ding{52} & \ding{52} & \ding{52} & & & \ding{52} & \ding{52} & 5 \\
  & Canon & \ding{52} & \ding{52} & \ding{52} & & & & \ding{52} & 4 \\
  & MICOSinging & \ding{52} & \ding{52} & & & & \ding{52} & \ding{52} & 4 \\
  & Clarinet & \ding{52} & \ding{52} & \ding{52} & & & \ding{52} & \ding{52} & 5 \\
  & Trumpet & \ding{52} & \ding{52} & \ding{52} & & & & & 3 \\
  & PianoSaxophone & \ding{52} & \ding{52} & \ding{52} & & & \ding{52} & \ding{52} & 5 \\
  & Chorus & \ding{52} & \ding{52} & \ding{52} & & & & \ding{52} & 4 \\
  & Studio & \ding{52} & \ding{52} & \ding{52} & \ding{52} & & & \ding{52} & 5 \\
  & Church & \ding{52} &\ding{52} & \ding{52} & & & & \ding{52} & 4 \\
  & Duet & \ding{52} & \ding{52} & & & & \ding{52} & \ding{52} & 4 \\
  & Blues & \ding{52} & \ding{52} & \ding{52} & & & & \ding{52} & 4 \\
  & Violins & \ding{52} & \ding{52} & \ding{52} & & \ding{52} & & \ding{52} & 5 \\
  & SingingDancing & \ding{52} & \ding{52} & \ding{52} & \ding{52} & & \ding{52} & \ding{52} & 6 \\
  \midrule
  \multirow{16}{*}{\begin{sideways}Miscellanea (16)\end{sideways}} &
  Beach & \ding{52} & \ding{52} & \ding{52} & \ding{52} & & & & 4 \\
  & BadmintonGym & \ding{52} & \ding{52} & \ding{52} & \ding{52} & & & & 4 \\
  & InVehicle & \ding{52} & \ding{52} & \ding{52} & & & \ding{52} & & 4 \\
  & Japanese & & & & \ding{52} & \ding{52} & \ding{52} & \ding{52} & 4 \\
  & Tennis & \ding{52} & \ding{52} & \ding{52} & \ding{52} & & \ding{52} & & 5 \\
  & Diesel & \ding{52} & \ding{52} & \ding{52} & & & \ding{52} & & 4 \\
  & Park & \ding{52} & \ding{52} & \ding{52} & \ding{52} & & & & 4 \\
  & Lion & & & \ding{52} & \ding{52} & & & & 2 \\
  & Carriage & \ding{52} & \ding{52} & \ding{52} & \ding{52} & \ding{52} & \ding{52} & & 6  \\
  & Platform  & \ding{52} & \ding{52} & \ding{52} & \ding{52} & & \ding{52} & & 5 \\
  & Dog & \ding{52} & & \ding{52} & \ding{52} & & \ding{52} & & 4 \\
  & RacingCar & \ding{52} & & \ding{52} & & & \ding{52} & \ding{52} & 4 \\
  & Train  & & \ding{52} & & \ding{52} & & \ding{52} & \ding{52} & 4 \\
  & Football & \ding{52} & \ding{52} & \ding{52} & \ding{52} & & & & 4 \\
  & ParkingLot & \ding{52} & \ding{52} & \ding{52} & \ding{52} & & \ding{52} & \ding{52} & 6 \\
  & Skiing & \ding{52} & \ding{52} & \ding{52} & \ding{52} & & \ding{52} & \ding{52} & 6 \\
  \midrule
  \\
  No. &  & 56 & 52 & 59 & 40 & 8 & 39 & 35 & 289 \\
  \bottomrule
  \end{tabular}
  %}
\end{table}

\begin{figure*}[t!]
	\centering
	\begin{overpic}[width=1\textwidth]{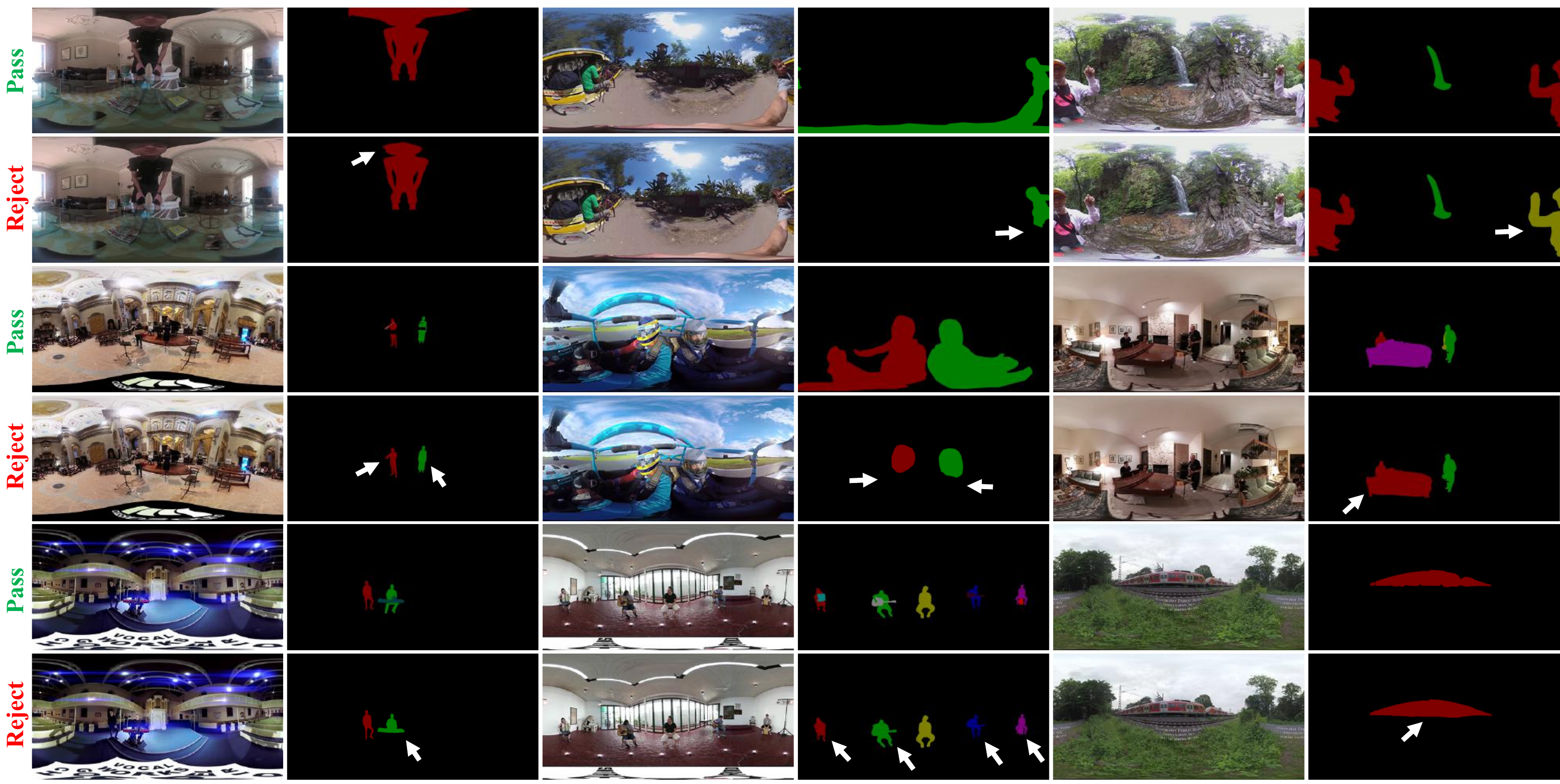}
    \end{overpic}
	\caption{Passed and rejected examples of annotation quality control.}
    \label{fig:Pass&Reject2}
\end{figure*}

\subsection{Dataset Features and Statistics}\label{sec:FeaStatis}

\noindent
\textbf{Attribute Distribution.}
The attributes summarize natural daily scenes viewed 
in omni-direction, also inspire model development for PI-SOD and PV-SOD.
As shown in \figref{fig:attributes_corr}, the seven proposed attributes are closely related to each other, representing challenging common scenarios. 

\noindent
\textbf{Equator Center Bias.}
\figref{fig:GTComparison} and \figref{fig:center_bias_class} visualize the global and super-class center bias~\cite{fan2018salient,fan2021salient} of \ourdataset.
Compared to 360-SOD and 360-SSOD, our dataset shows stronger center bias as it is the only one with salient objects annotated according to participants' eye fixations, with the starting point set to the center of each video display during the subjective experiments.
It has been broadly proved that photographers tend to capture the main content of a 360° video at the equator center, and that users also usually pay more attention to such an area during watching~\cite{xu2020state,salient360vid, fan2019survey}. Our \ourdataset~is hence more able to reflect real-world viewing behaviors compared to the 360-SOD \cite{li2020distortion} and 360-SSOD \cite{ma2020stage}.

\noindent
\textbf{Instance Size.}
Following~\cite{fan2018salient}, we compute the normalized instances' size of our \ourdataset. The size distribution ranges from 0.03\% $\sim$ 23.00\%, covering very small objects.

\noindent
\textbf{Quality Control.}
High-quality annotation is one of the most important aspects of training for learning-based models. As illustrated in \figref{fig:Pass&Reject2}, we carefully conduct three-fold cross-validation to ensure the annotation quality.

\section{Empirical Studies}\label{sec:benchmark}

\subsection{Settings}\label{sec:experiment_setting}

\noindent
\textbf{Dataset Splits.}
All 67 videos are split into separate training and test sets with a random selection strategy, in a ratio of about 6:4. Therefore, we reach a unique split of 40 training and 27 test videos (5,796/4,669 key frames respectively), with corresponding per-pixel instance-/object-level GTs. The testing set is further divided into test0/test1/test2 with 6/6/15 videos, respectively, according to super-class labels (\ie, Miscellanea/Music/Speaking).%保留大写

\noindent
\textbf{Metrics.} 
We apply three widely used SOD metrics to quantitatively compare the SOTA I-SOD/V-SOD models. 
These metrics include structural measure (S-Measure, $S_\alpha$)~\cite{Fan2017Smeasure,fan2021structure}, maximum enhanced-alignment measure (E-Measure, $E_\phi$)~\cite{fan2021cognitive,fan2018enhanced} and mean absolute error (MAE)~\cite{MAE}. 
The MAE \cite{MAE} focuses on the local (per-pixel) match between ground truth and prediction, while S-Measure ($S_\alpha$) \cite{Fan2017Smeasure} pays attention to the object structure similarities. Besides, E-Measure ($E_\phi$) \cite{fan2018enhanced} considers both the local and global information.
\par
\noindent
\textbf{MAE} computes the mean absolute error between the ground truth $G \in \{0, 1\}$ and a normalized predicted saliency map $P \in [0, 1]$, i.e.,
\begin{equation}\label{equ:mae}
   MAE = \frac{1}{W\times{H}}\sum_{i=1}^{W}\sum_{j=1}^{H}\mid G(i, j) - P(i, j)\mid,
\end{equation}
where $H$ and $W$ denotes height and width, respectively.
\par
\noindent
\textbf{S-Measure} evaluates the structure similarities between salient objects in GT foreground maps and predicted saliency maps:
\begin{equation}\label{equ:sm}
   S_\alpha = \alpha \times S_{o} + (1 - \alpha) \times S_{r}.
\end{equation}
where $S_{o}$ and $S_{r}$ denotes the object-/region-based structure similarities, respectively. $\alpha \in [0,1]$ is set as 0.5 so that equal weights are assigned to both the object-level and region-level assessments \cite{Fan2017Smeasure}.
\par
\noindent
\textbf{E-Measure} is a cognitive vision-inspired metric to evaluate both the local and global similarities between two binary maps. Specifically, it is defined as:
\begin{equation}\label{equ:em}
E_{\phi}=\frac{1}{W\times H}\sum_{i=1}^W\sum_{j=1}^H\phi_{s}(i,j),
\end{equation} 
where $\phi_{s}(i,j)$ represents the enhanced alignment matrix \cite{fan2018enhanced}. 

%In this paper, we report maximum E-Measure scores for the quantitative comparison between all \OurTotalBaselines~baselines, with E-Measure curves shown in Figure \ref{fig:curves_class} and Figure \ref{fig:curves_attr}.

\begin{table*}[t!]
  \centering
  \footnotesize
  \renewcommand{\arraystretch}{1.0}
  \setlength\tabcolsep{5.5pt}
  \caption{
   Performance comparison of 7/3 state-of-the-art conventional I-SOD/V-SOD methods and one PI-SOD method~\cite{huang2020fanet} over \ourdataset. %$S_\alpha$ = S-measure ($\alpha$=0.5 \cite{Fan2017Smeasure}), $F_\beta$ = maximum F-measure ($\beta$=0.3) \cite{Fmeasure}, $E_\phi$ = maximum E-measure \cite{fan2018enhanced}, $\mathcal{M}$ = mean absolute error \cite{MAE}.
   $\uparrow$/$\downarrow$ denotes a larger/smaller value is better. 
   Best result of each column is \textbf{bolded}. 
   %Evaluation code: \url{https://github.com/zzhanghub/eval-co-sod}
   }
   \label{tab:QuantityComparison}
  \begin{tabular}{c|r|r||ccc|ccc|ccc}
   \toprule
 %  &\multirow{2}{*}{Methods} & \multicolumn{4}{c||}{Test0 (Miscellanea)} & \multicolumn{4}{c||}{Test1 (Music)} & \multicolumn{4}{c}{Test2 (Speaking)}  
    \multirow{2}{*}{Type} & \multirow{2}{*}{Publication} &\multirow{2}{*}{Methods} & \multicolumn{3}{c|}{Miscellanea (Test0)} & \multicolumn{3}{c|}{Music (Test1)} & \multicolumn{3}{c}{Speaking (Test2)}  
  \\
  \cline{4-12}
  % && $F_{\beta}~\uparrow~~$ & $S_{\alpha}~\uparrow~~$ & $E_\phi~\uparrow~~$ & $\mathcal{M}~\downarrow~~$  
  %  & $F_{\beta}~\uparrow~~$ & $S_{\alpha}~\uparrow~~$ & $E_\phi~\uparrow~~$ & $\mathcal{M}~\downarrow~~$ 
  %  & $F_{\beta}~\uparrow~~$ & $S_{\alpha}~\uparrow~~$ & $E_\phi~\uparrow~~$ & $\mathcal{M}~\downarrow~~$ 
    & &&  $S_{\alpha}~\uparrow~~$ & $E_\phi~\uparrow~~$ & $\mathcal{M}~\downarrow~~$  
    & $S_{\alpha}~\uparrow~~$ & $E_\phi~\uparrow~~$ & $\mathcal{M}~\downarrow~~$ 
    &  $S_{\alpha}~\uparrow~~$ & $E_\phi~\uparrow~~$ & $\mathcal{M}~\downarrow~~$ 
  \\
  \hline
   \multirow{7}{*}{I-SOD}
  &CVPR'19 & CPD \cite{CPD} & 0.654 & 0.584 & 0.035 & 0.608 & 0.823 & 0.018  & 0.588 & 0.756 & 0.026 \\
  &ICCV'19 & SCRN \cite{SCRN}  & 0.665 & 0.564 & 0.046  & 0.683 & 0.841 & 0.023 & \textbf{0.636} & 0.739 & 0.034 \\
  &AAAI'20 & F3Net \cite{F3Net}  & 0.655 & 0.557 & 0.040& 0.662 & 0.801 & 0.021 & 0.626 & 0.716 & 0.027 \\
  &CVPR'20 & MINet \cite{MINet} & 0.650 & 0.557 & 0.050  & 0.670 & 0.789 & 0.020 & 0.590 & 0.680 & 0.053 \\
  &CVPR'20 & LDF \cite{CVPR2020LDF}& 0.663 & 0.557 & 0.044 & 0.671 & 0.828 & 0.023 & 0.625 & 0.761 & 0.037 \\
  &ECCV'20 & CSF \cite{SOD100K}  & 0.652 & 0.575 & 0.033 & 0.665 & 0.833 & 0.018 & \textbf{0.636} & \textbf{0.791} & 0.026 \\
  &ECCV'20 & GateNet \cite{GateNet} & \textbf{0.677} & \textbf{0.596} & 0.044 & 0.673 & \textbf{0.852} & 0.018 & 0.633 & 0.739 & 0.034 
  \\
  \hline
  \multirow{3}{*}{V-SOD}
  &CVPR'19 & COSNet \cite{COSNet} & 0.610 & 0.535 & 0.031 & 0.577 & 0.825 & \textbf{0.016} & 0.572 & 0.722 & \textbf{0.023} \\
  &ICCV'19 & RCRNet \cite{RCRNet}  & 0.661 & 0.576 & 0.034 & \textbf{0.695} & 0.839 & 0.019 & 0.632 & 0.775 & 0.030 \\
  &AAAI'20 & PCSA \cite{gu2020PCSA} & 0.602 & 0.549 & 0.034  & 0.655 & 0.764 & 0.021& 0.572 & 0.679 & 0.026
  \\
  \hline
%   \multirow{1}{*}{PI-SOD}
  PI-SOD
  %& FANet \cite{huang2020fanet} & 0.192 & 0.610 & 0.513 & 0.030 & 0.408 & 0.646 & 0.814 & 0.018 & 0.263 & 0.566 & 0.696 & 0.027
  & SPL'20 & FANet \cite{huang2020fanet} & 0.610 & 0.513 & \textbf{0.030} & 0.646 & 0.814 & 0.018 & 0.566 & 0.696 & 0.027
  \\
  \bottomrule
  \end{tabular}
  %}
\end{table*}

\begin{table*}[t!]
  \centering
  \renewcommand{\arraystretch}{1.0}
  \setlength\tabcolsep{0.1pt}
  \footnotesize
  %\resizebox{0.99\textwidth}{!}{
  \caption{
   Performance comparison of 7/3/1 state-of-the-art I-SOD/V-SOD/PI-SOD methods based on each of the attributes. %of our \textit{\ourdataset}. 
   %$S_\alpha$ = S-measure ($\alpha$=0.5 \cite{Fan2017Smeasure}), $F_\beta$ = maximum F-measure ($\beta$=0.3) \cite{Fmeasure}, $E_\phi$ = maximum E-measure \cite{fan2018enhanced}, $\mathcal{M}$ = mean absolute error \cite{MAE}. %$\uparrow$/$\downarrow$ denotes a larger/smaller value is better. 
   %Best result of each column is \textbf{bolded}. 
   %Evaluation code: \url{https://github.com/zzhanghub/eval-co-sod}
   }\label{tab:QuantityComparisonAttr}
  \begin{tabular}{l|r||ccccccc|ccc|c}
   \toprule
   \multirow{2}{*}{Attr.}&\multirow{2}{*}{Metrics} & \multicolumn{7}{c|}{I-SOD} & \multicolumn{3}{c|}{V-SOD} & \multicolumn{1}{c}{PI-SOD}
  \\
  \cline{3-13}
   && CPD \cite{CPD}~ & SCRN \cite{SCRN}~ & F3Net \cite{F3Net}~ & MINet \cite{MINet}~ & LDF \cite{CVPR2020LDF}~ & CSF \cite{SOD100K}~ & GateNet \cite{GateNet}~ & COSNet \cite{COSNet}~ & RCRNet \cite{RCRNet}~ & PCSA \cite{gu2020PCSA}~ & FANet \cite{huang2020fanet}
   \\
  \hline
  \multirow{3}{*}{MO} 
 % & $F_{\beta}~\uparrow$ & 0.285 & \textbf{0.354} & 0.315 & 0.314 & 0.321 & 0.334 & 0.336 & 0.293 & 0.346 & 0.262 \\
                      & $S_{\alpha}~\uparrow$ & 0.610 & 0.657 & 0.644 & 0.624 & 0.648 & 0.649 & 0.653 & 0.588 & \textbf{0.661} & 0.606 & 0.605 \\
                      & $E_{\phi}~\uparrow$ & 0.741 & 0.740 & 0.702 & 0.691 & 0.742 & \textbf{0.752} & 0.733 & 0.722 & 0.746 & 0.681 & 0.695\\
                      & $\mathcal{M}~\downarrow$ & 0.027 & 0.034 & 0.030 & 0.045 & 0.033 & 0.027 & 0.034 & \textbf{0.024} & 0.029 & 0.027 & 0.025 \\ 
  \hline
  \multirow{3}{*}{OC}
  %& $F_{\beta}~\uparrow$ & 0.312 & \textbf{0.376} & 0.344 & 0.334 & 0.346 & 0.356 & 0.359 & 0.310 & 0.358 & 0.281 \\
                      & $S_{\alpha}~\uparrow$ & 0.606 & \textbf{0.655} & 0.641 & 0.619 & 0.645 & 0.645 & 0.650 & 0.577 & 0.652 & 0.599 & 0.593 \\
                      & $E_{\phi}~\uparrow$ & 0.772 & 0.768 & 0.725 & 0.699 & 0.771 & \textbf{0.780} & 0.755 & 0.744 & 0.763 & 0.704 & 0.720 \\
                      & $\mathcal{M}~\downarrow$ & 0.023 & 0.029 & 0.026 & 0.043 & 0.028 & 0.023 & 0.030 & \textbf{0.020} & 0.025 & 0.024 & 0.022 \\ 
  \hline
  \multirow{3}{*}{LS} 
  %& $F_{\beta}~\uparrow$ & 0.272 & \textbf{0.336} & 0.301 & 0.298 & 0.299 & 0.315 & 0.320 & 0.273 & 0.320 & 0.252 \\
                      & $S_{\alpha}~\uparrow$ & 0.605 & 0.649 & 0.639 & 0.618 & 0.637 & 0.644 & 0.647 & 0.585 & \textbf{0.650} & 0.609 & 0.598 \\
                      & $E_{\phi}~\uparrow$ & 0.721 & 0.723 & 0.693 & 0.665 & 0.719 & \textbf{0.740} & 0.715 & 0.697 & 0.723 & 0.674 & 0.669 \\
                      & $\mathcal{M}~\downarrow$ & 0.025 & 0.034 & 0.028 & 0.045 & 0.037 & 0.025 & 0.033 & \textbf{0.022} & 0.029 & 0.026 & 0.025 \\ 
  \hline
  \multirow{3}{*}{MB} 
  %& $F_{\beta}~\uparrow$ & 0.332 & \textbf{0.374} & 0.313 & 0.332 & 0.342 & 0.342 & 0.346 & 0.317 & 0.343 & 0.273 \\
                      & $S_{\alpha}~\uparrow$ & 0.622 & \textbf{0.651} & 0.630 & 0.620 & 0.646 & 0.638 & 0.645 & 0.582 & 0.642 & 0.586 & 0.587 \\
                      & $E_{\phi}~\uparrow$ & 0.728 & 0.718 & 0.692 & 0.675 & 0.717 & \textbf{0.749} & 0.701 & 0.709 & 0.734 & 0.702 & 0.688 \\
                      & $\mathcal{M}~\downarrow$ & 0.021 & 0.029 & 0.027 & 0.047 & 0.029 & 0.021 & 0.030 & \textbf{0.019} & 0.024 & 0.022 & 0.020 \\ 
  \hline
  \multirow{3}{*}{OV}
  %& $F_{\beta}~\uparrow$ & 0.368 & \textbf{0.469} & 0.169 & 0.337 & 0.321 & 0.349 & 0.357 & 0.294 & 0.294 & 0.273 \\
                      & $S_{\alpha}~\uparrow$ & 0.634 & \textbf{0.661} & 0.568 & 0.633 & 0.636 & 0.636 & 0.639 & 0.582 & 0.630 & 0.600 & 0.611 \\
                      & $E_{\phi}~\uparrow$ & 0.844 & 0.786 & 0.571 & 0.711 & \textbf{0.854} & 0.837 & 0.841 & 0.817 & 0.848 & 0.700 & 0.820 \\
                      & $\mathcal{M}~\downarrow$ & \textbf{0.018} & 0.021 & 0.029 & 0.038 & 0.039 & 0.021 & 0.025 & 0.021 & 0.029 & 0.021 & \textbf{0.018} \\ 
  \hline
   \multirow{3}{*}{GD} 
   %& $F_{\beta}~\uparrow$ & 0.308 & \textbf{0.355} & 0.290 & 0.309 & 0.331 & 0.336 & 0.330 & 0.298 & 0.325 & 0.250 \\
                      & $S_{\alpha}~\uparrow$ & 0.630 & \textbf{0.662} & 0.639 & 0.633 & 0.659 & 0.646 & 0.658 & 0.588 & 0.651 & 0.578 & 0.599 \\
                      & $E_{\phi}~\uparrow$ & 0.680 & 0.690 & 0.641 & 0.666 & 0.672 & \textbf{0.695} & 0.684 & 0.662 & \textbf{0.695} & 0.655 & 0.657 \\
                      & $\mathcal{M}~\downarrow$ & 0.037 & 0.042 & 0.040 & 0.045 & 0.043 & 0.035 & 0.042 & \textbf{0.032} & 0.037 & 0.036 & 0.034 \\ 
  \hline
   \multirow{3}{*}{CS}
   %& $F_{\beta}~\uparrow$ & 0.314 & \textbf{0.382} & 0.355 & 0.355 & 0.355 & 0.371 & 0.372 & 0.335 & 0.380 & 0.295 \\
                      & $S_{\alpha}~\uparrow$ & 0.625 & \textbf{0.680} & 0.667 & 0.654 & 0.664 & 0.670 & 0.676 & 0.592 & \textbf{0.680} & 0.620 & 0.616 \\
                      & $E_{\phi}~\uparrow$ & 0.748 & 0.759 & 0.712 & 0.718 & 0.747 & 0.745 & \textbf{0.762} & 0.722 & 0.736 & 0.689 & 0.710 \\
                      & $\mathcal{M}~\downarrow$ & 0.029 & 0.035 & 0.031 & 0.035 & 0.034 & 0.028 & 0.033 & \textbf{0.026} & 0.030 & 0.029 & 0.028 \\ 
  \bottomrule
  \end{tabular}
\end{table*}

% S-Measure 均值： 0.631(MO) 0.626(OC) 0.626(LS) 0.623(MB)	0.621(OV) 0.631(GD) 0.649(CS)

\noindent
\textbf{Training Protocols.} 
To provide a comprehensive benchmark, we collect the released codes of 10 SOTA I-SOD/V-SOD methods and one PI-SOD model, re-train these models with the training set of \ourdataset~and the widely used I-SOD training set, DUTS-train \cite{DUTS} (except for FANet \cite{huang2020fanet}, which is designed for ER images only). 
The selected baselines (CPD\cite{CPD}, SCRN\cite{SCRN}, F3Net\cite{F3Net}, MINet\cite{MINet}, LDF\cite{CVPR2020LDF}, CSF+Res2Net\cite{SOD100K}, GateNet \cite{GateNet}, RCRNet\cite{RCRNet}, COSNet\cite{COSNet} and PCSA \cite{gu2020PCSA}) meet the following criteria: i) classical architectures, ii) recently published and open-sourced, iii) achieve SOTA performance on existing I-SOD/V-SOD benchmarks.
Note that all the baselines are trained with the recommended parameter settings.

\section{Discussion}
From the benchmark results, we observe that, generally, the I-SOD models gain comparable performance to their V-SOD counterparts.
One possible reason is that, since the annotations of our \ourdataset~are only based on key frames, the relatively sparse spatiotemporal information may prevent the V-SOD models from acquiring full performance. In contrast, the visual cues are easily learned by the I-SOD model with such sparsely labeled data. 
%Next, we discuss seven open issues, that have not been fully addressed by the existing PV-SOD field and should be explored in the future.

\begin{figure*}[t!]
	\centering
	\begin{overpic}[width=.98\textwidth]{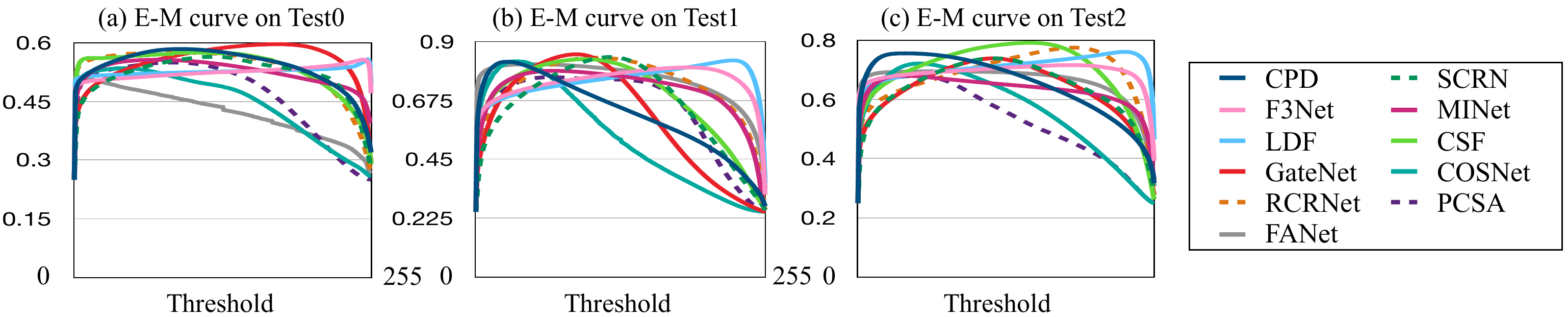}
    \end{overpic}
	\caption{E-Measure (E-M) curves of all baselines upon our \textit{\ourdataset}.
	}\label{fig:curves_class}
\end{figure*}

\noindent
\textbf{Overall Performance.}
From the evaluation in \tabref{tab:QuantityComparison}, we observe that, in most cases, the I-SOD methods (\eg, GateNet, and CSF) achieve better performance than the V-SOD (\eg, COSNet, RCRNet, and PCSA) and PI-SOD models.
For specific scenes (\eg, speaking), however, SCRN obtains a very competitive performance to GateNet, while performs worse than CSF. The E-Measure performances are shown in \figref{fig:curves_class}.

\begin{figure*}[t!]
	\centering
	\begin{overpic}[width=.98\textwidth]{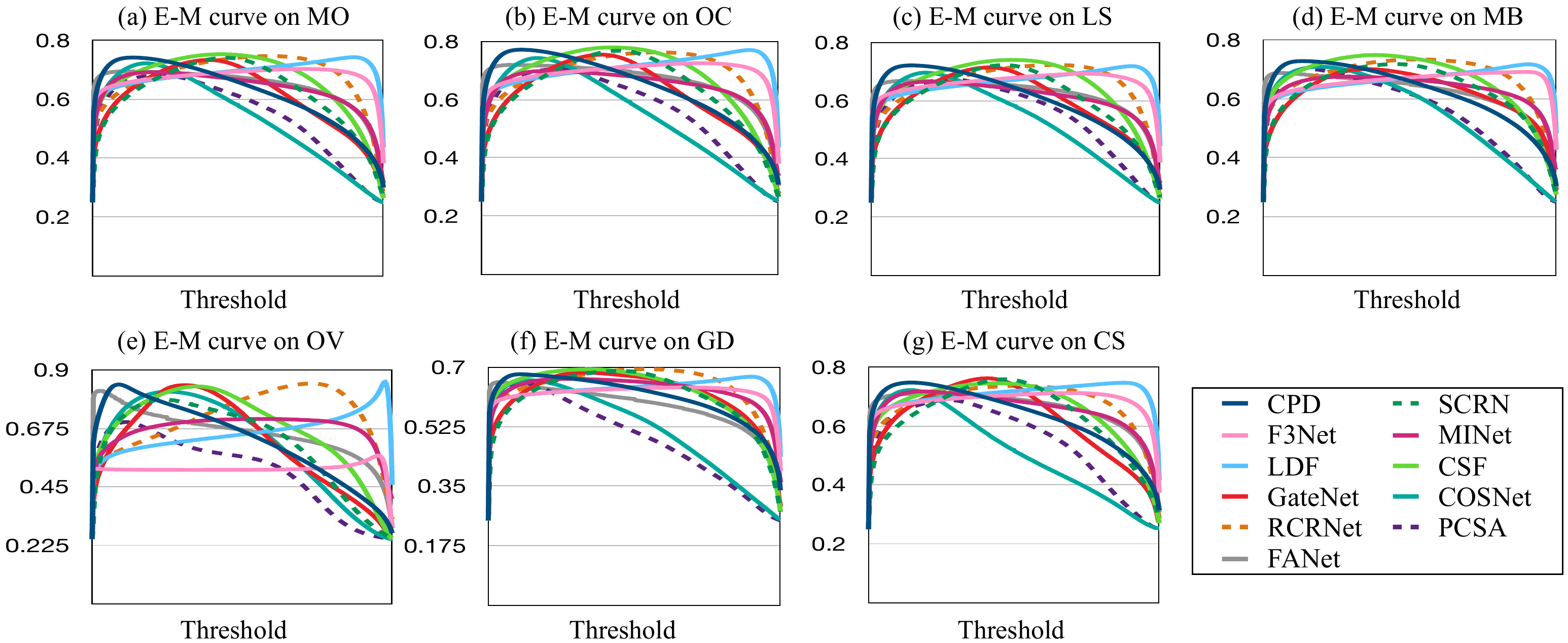}
    \end{overpic}
	\caption{Attribute-based E-Measure (E-M) curves of all baselines upon our \textit{\ourdataset}.
	}\label{fig:curves_attr}
\end{figure*}

\noindent
\textbf{Attribute Performance.}
To provide deeper insights for the challenging cases, we report the performances of  all \OurTotalBaselines~baselines on our seven attributes. A detailed attributes-based E-Measure is shown in \figref{fig:curves_attr}. As shown in \tabref{tab:QuantityComparisonAttr}, the average $S_\alpha$ score among the models for the different attributes are: 0.631 (MO), 0.626 (OC), 0.626 (LS), 0.623 (MB), 0.621 (OV), 0.631 (GD) and 0.649 (CS). 
\textit{Out-of-view} (OV) is the most challenging attribute as the objects usually appear in the corner of the ER images. 
Besides, the scores on all attributes are less than 0.65, demonstrating the strong challenge of our \ourdataset and leaving large room for future improvement.

\begin{figure}[b!]
	\centering
    \begin{overpic}[width=.98\columnwidth]{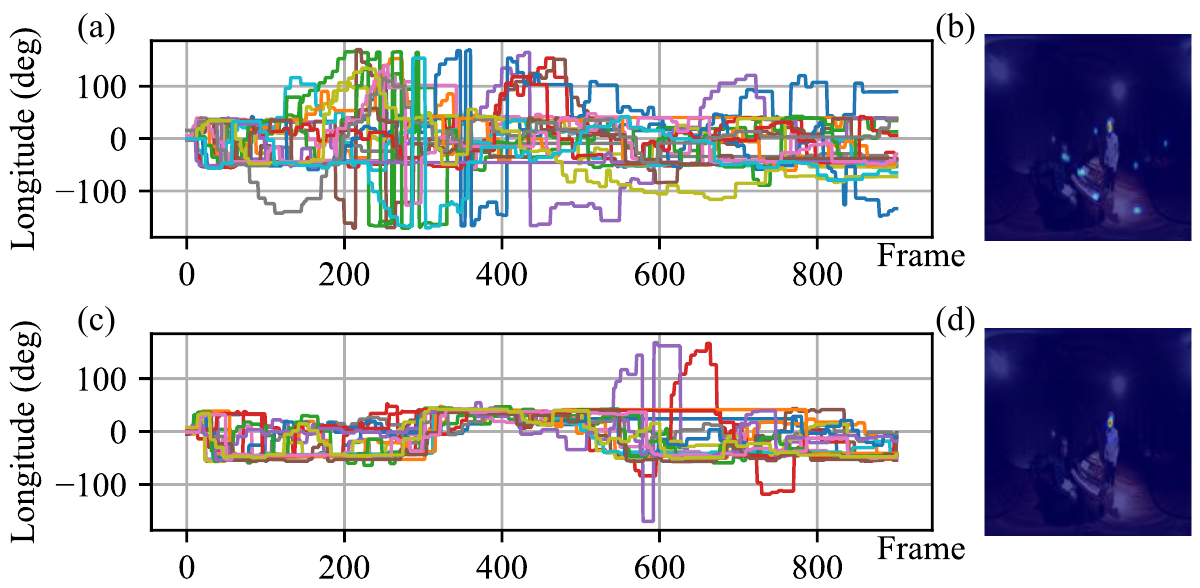}
    \end{overpic}
   
    \caption{
    Eye fixation distributions of all participants watching \textit{PianoConvo} sequence without (a)/with (c) audio. Corresponding fixations are overlaid in (b) and (d), respectively.
    }\label{fig:fixa_distri}
\end{figure}

\noindent
\textbf{General Attributes.} As shown in Table \ref{tab:attributes_details}, every collected video owns at least one general attributes (\ie, \textit{multiple objects (MO)}, \textit{occlusions (OC)}, \textit{motion blur (MB)} and \textit{low space (LS)}), indicating that our \ourdataset~contains the main challenges faced in so many computer vision fields, such as object detection, video object segmentation, etc. It is worth mentioning that, as the 360° image captures a wide field-of-view (FoV) with the range of 360°$\times$180°, salient objects far from the panoramic camera may perform extremely small sizes, making the \textit{LS} a more challenging situation within \ourtask.

\begin{figure}[t!]
	\centering
	\begin{overpic}[width=\columnwidth]{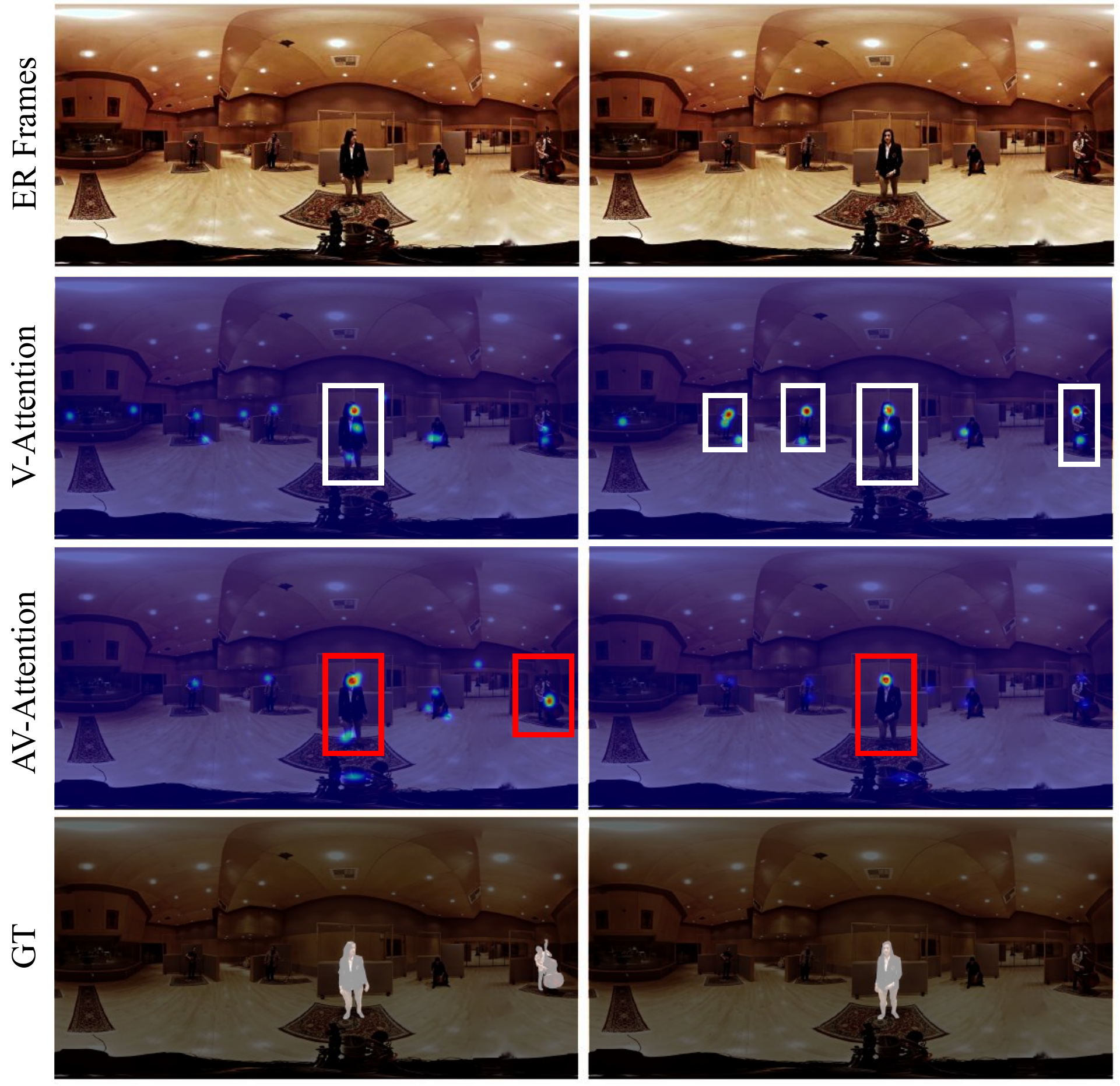}
    \end{overpic}
	\caption{
	Visual comparison of visual and audio-visual attention. The objects with high saliency values ($\geq50\%$) are marked with white/red bounding boxes. Zoom in for details.	
	}\label{fig:AVAttenGT}
\end{figure}

\noindent
\textbf{Eye Fixations With/Without Audio.}
\figref{fig:fixa_distri} shows an example of 20 participants watching videos without and with audio in longitude, respectively. We find that the eye fixation recordings with audio are highly consistent across subjects, while the recordings without audio are not. 
%This finding is similar with the conclusion that ``HM positions are highly consistent across humans.''~\cite{pvshm}. 
\figref{fig:AVAttenGT} highlights this finding by vividly showing a significant divergence between human attentions (converging/scattering) with and without the guidance of audio information, respectively. Both the quantitative (\figref{fig:fixa_distri}) and qualitative (\figref{fig:AVAttenGT}) experimental results indicate that human attention highly depends on a co-guidance of audio-visual information.

\begin{figure*}[t!]
	\centering
	\begin{overpic}[width=\textwidth]{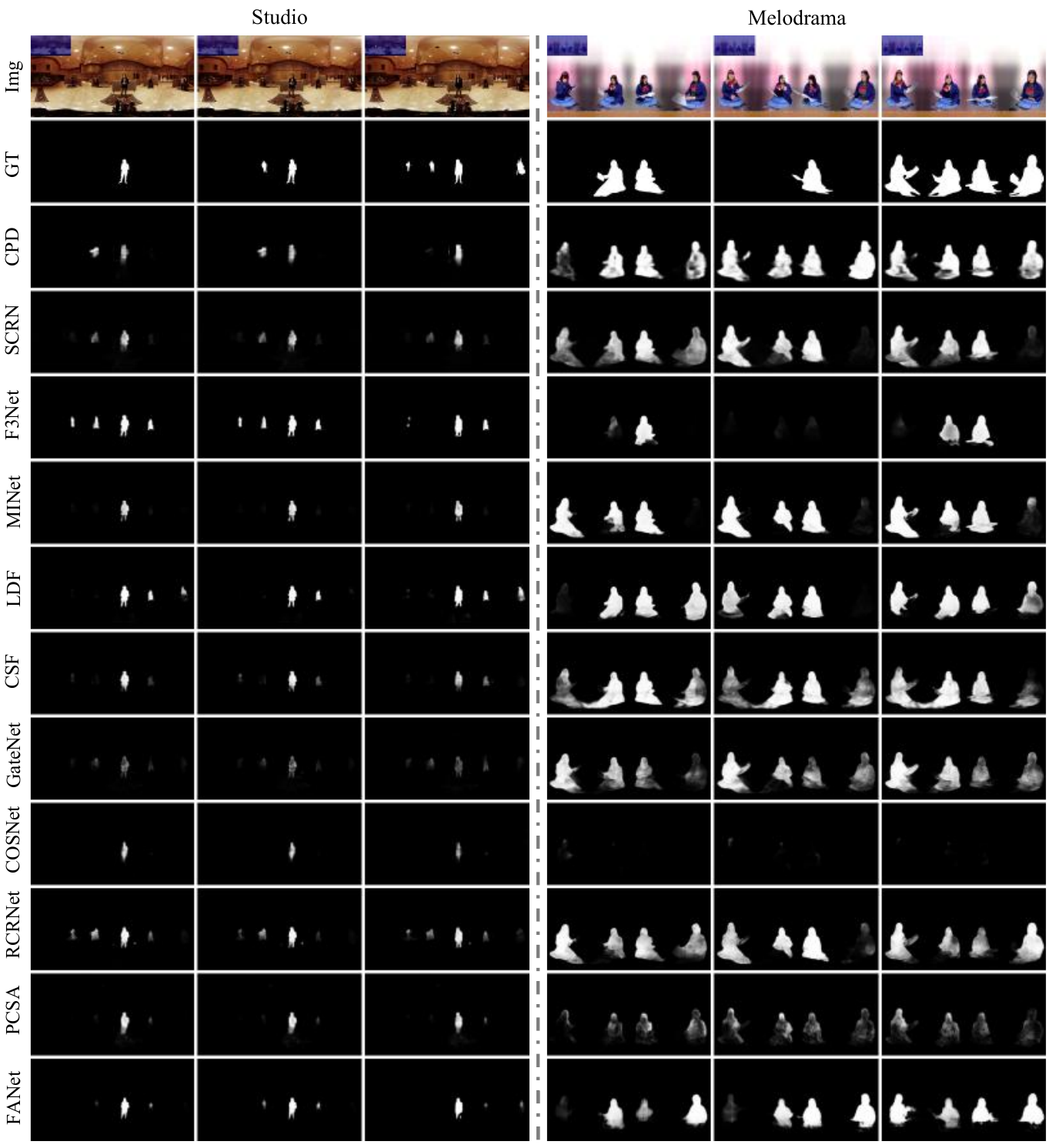}
    \end{overpic}
	\caption{An illustration of the unique audio-visual attribute, competing sounds. Img = image. GT = ground truth.}
    \label{fig:visual_cs}
\end{figure*}

\noindent
\textbf{Audio-Induced Attributes.} Creating realistic VR experiences requires 360° videos to be captured with their surrounding visual and audio stimuli. The audio cue plays a significant role in informing the viewers about the location of sounding salient objects in the 360° environment \cite{vasudevan2020semantic}, providing an immersive multimedia experience. 
%According to the neuron-science and psychological studies~\cite{HM_sound_loco, HM_vestibular} about the effects of rotational head movement in improving the accuracy of sound localization, audio signal is considered as a powerful way of directing viewers’ attention~\cite{Samuel_Beckett_in_vr}. 
%
%However, existing SOD methods ignore the audio cue and thus failing to detect the correct salient objects annotated based on audio-induced eye fixations (Figure \ref{fig:visual_cs}).
\yz{As shown in Figure \ref{fig:visual_cs}, the ground truths (GTs) of our videos (especially for those attached with attribute-\emph{competing sounds}) may be largely influenced by the audio cues (\eg, silent visual salient objects are sometimes regarded as non-salient objects in our case, for they are unable to gain a majority of the observers' attention in an acoustic environment), thus being very different when compared with the GTs of existing mute-modal V-SOD datasets where visual cues are the only evidence.}
As shown in Figure \ref{fig:visual_cs}, most of the baselines tend to detect all visual salient objects while ignore the audio-visual attention shifts among the key frames. It is necessary for future works to model both the visual and audio cue for a better performance on our \ourdataset.

\begin{figure}[t!]
	\centering
	\begin{overpic}[width=.75\columnwidth]{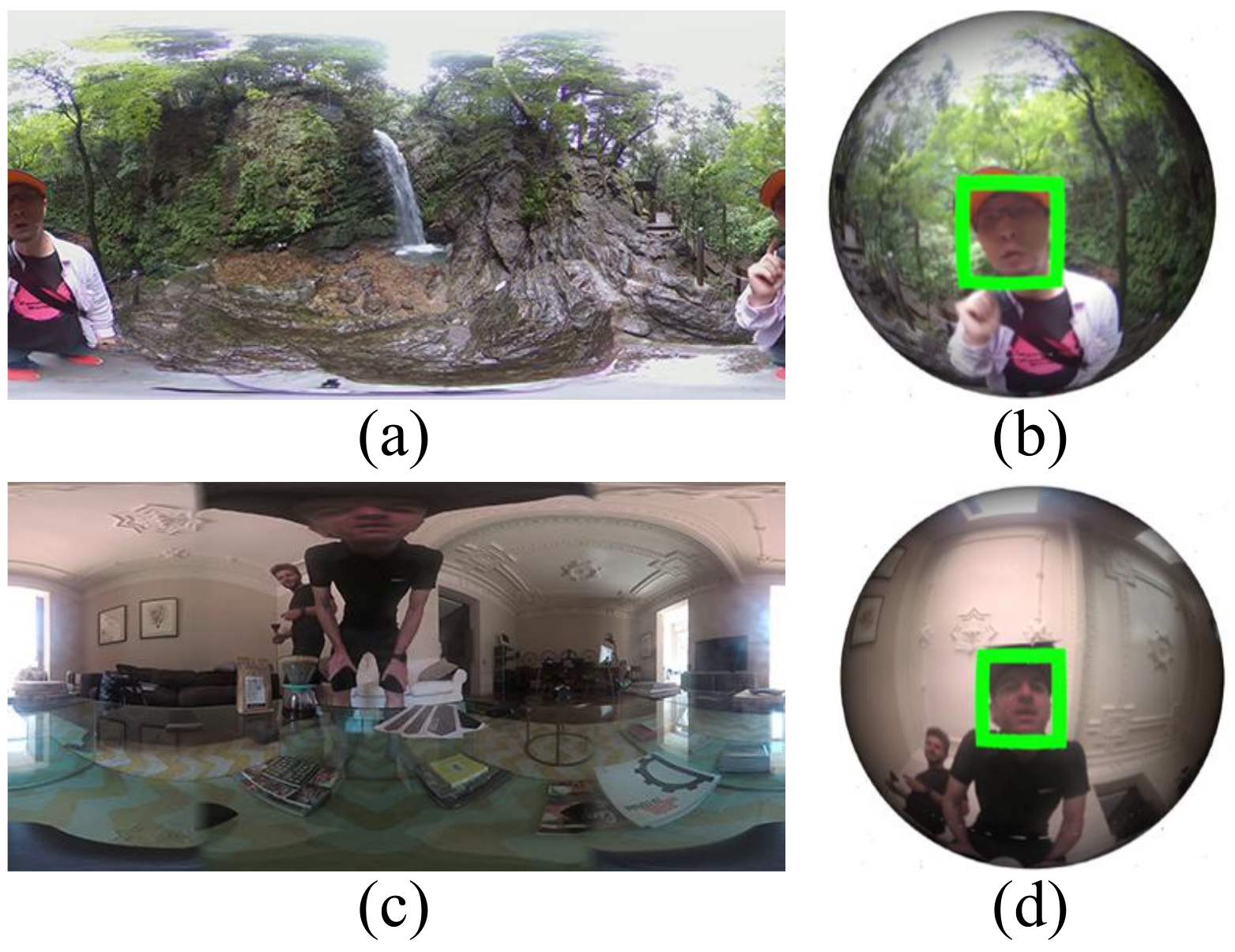}
    \end{overpic}
	\caption{An illustration of typical geometrical attributes. (a) Out-of-view (OV) in equirectangular (ER) image. (b) OV in spherical field-of-view (FoV). (c) Geometrical distortion (GD) in ER image. (d) GD in spherical FoV.}\label{fig:visual_ga}
\end{figure}

\noindent
\textbf{360° Geometrical Attributes.} The 360° image captures a scene covering omnidirectional (360°$\times$180°) spatial information, thus including more comprehensive object structures when compared to 2D image, which owns a FoV with limited range. For instance, as shown in Figure \ref{fig:visual_ga}, the \textit{out-of-view (OV)} object in equirectangular (ER) image has a well shape when re-projected to a specific FoV on sphere. However, \textit{OV} objects in 2D images or videos permanently lose spatial information due to the limited viewing range of the normal cameras. \textit{Geometrical distortion (GD)} is the other important attribute of 360° images under ER projection (Figure \ref{fig:visual_ga} (c)), which is largely alleviated in spherical FoV (Figure \ref{fig:visual_ga} (d)). As there is no perfect 2D (planar) representation for 360° images, a trade-off between geometrical distorted extent and spatial information retention will always exist. Future methods may take advantage of multiple projection methods for 
%\yz{an} 
improved performance when conducting \ourtask.

%that human attention (third row) mainly attracted by the middle and right person when providing the video with audio, while human attention records are scattered everywhere when providing the video without audio.

\noindent
\textbf{Small Objects.}
We define small objects \textit{LS} (\tabref{tab:Attributes}) as those that occupy an area smaller than 0.5\% of the whole image. \textit{LS}, as one of the well-known challenges in the image segmentation task, is still not completely solved. As stated in \secref{sec:FeaStatis}, due to the wide FoV range of 360°$\times$180°, the smallest object in our \ourdataset~only occupies 0.03\% of the ER image, making it more challenging. As for the attribute-based performance, we observe that models under this situation achieve a comparable performance to the others. However, considering that the applied metrics may be biased toward the true negative, the true positive score tends to be worse than the existing shown performance.

\noindent
\textbf{Novel Metric.}
In this benchmark, we only introduce widely used metrics for I-SOD. However, PV-SOD involves context (\eg, audio, spatial and temporal) relationship between salient/non-salient objects, which is quite important for PV-SOD assessment. Thus, designing a more suitable evaluation metric for PV-SOD is an interesting and open issue. 

\noindent
\textbf{Future Directions.}
Currently, we only focus on the object-level task. However, the instance-level task is more difficult and may be suitable for many image-editing applications. 
In addition, as described in~\cite{fan2018salient}, studying non-salient objects will provide rich context for reasoning the salient objects in a scene.
\yz{Besides}, in this study, we only provide sparse annotations for the proposed dataset. However, dense annotations like those given in DAVIS~\cite{DAVIS} dataset can provide more valuable information (\eg, sequence-to-sequence modeling or audio-visual matching) for both traditional I-SOD and PV-SOD models. \yz{Finally, based on the findings of significant differences between visual and audio-visual saliency attributes as illustrated in ``Audio-Induced Attributes'' of Section Discussion, we find that an absence of audio cues may significantly limit current SOD models from acquiring their best performance on \ourdataset. Future researches may focus more on utilizing the audio cues for better object-level saliency detection in panoramic scenes.}

\section{Conclusion}\label{sec:conclusion}
We have proposed \ourdataset, the first large-scale dataset for the \ourtask~task. Compared with the traditional SOD task, 
\ourtask~is more challenging. 
The hierarchical annotations enable \ourdataset~to easily be extended to different-level tasks, such as weakly supervised learning, multi-modality learning, and head movement/fixation prediction. 
In addition, we provide several empirical rules for creating high-quality datasets. We have further investigated 11 cutting-edge methods at both the overall and attribute levels. The obtained findings indicate that \ourtask~is far from being solved. We hope that our studies will facilitate SOD research towards panoramic videos and thus inspiring more novel ideas for AR/VR applications.

\begin{table*}[t!]
  \centering
  \renewcommand{\arraystretch}{0.9}
  \setlength\tabcolsep{1.2pt}
   \caption{
   Sequence performance comparison of 7/3/1 SOTA I-SOD/V-SOD/PI-SOD methods. Sp. = Speaking. Mu. = Music.
   }\label{tab:SeqQua_1}
  \resizebox{0.99\textwidth}{!}{
  \begin{tabular}{l|r||ccccccc|ccc|c}
   \toprule
   \multirow{2}{*}{Super-class/Sequence}&\multirow{2}{*}{Metrics} & \multicolumn{7}{c|}{I-SOD} & \multicolumn{3}{c|}{V-SOD} & \multicolumn{1}{c}{PI-SOD}
  \\
  \cline{3-13}
   && CPD \cite{CPD}~ & SCRN \cite{SCRN}~ & F3Net \cite{F3Net}~ & MINet \cite{MINet}~ & LDF \cite{CVPR2020LDF}~ & CSF \cite{SOD100K}~ & GateNet \cite{GateNet}~ & COSNet \cite{COSNet}~ & RCRNet \cite{RCRNet}~ & PCSA \cite{gu2020PCSA}~ & FANet \cite{huang2020fanet}
   \\
  \hline
  \multirow{4}{*}{Sp./Debate} 
    & $S_{\alpha}~\uparrow$ & 0.547 & 0.620 & 0.605 & 0.553 & 0.566 & 0.576 & \textbf{0.628} & 0.514 & 0.559 & 0.571 & 0.557 \\
    & max $E_{\phi}~\uparrow$ & 0.764 & \textbf{0.855} & 0.854 & 0.800 & 0.842 & 0.853 & 0.849 & 0.844 & 0.843 & 0.836 & 0.755  \\
    & mean $E_{\phi}~\uparrow$ & 0.600 & 0.752 & 0.818 & 0.592 & \textbf{0.829} & 0.802 & 0.768 & 0.410 & 0.809 & 0.605 & 0.702  \\
    & $\mathcal{M}~\downarrow$ & 0.014 & 0.016 & 0.014 & 0.012 & 0.012 & 0.013 & 0.015 & \textbf{0.009} & 0.013 & 0.012 & 0.015 \\ 
  \hline
  \multirow{4}{*}{Sp./BadmintonConvo} 
    & $S_{\alpha}~\uparrow$ & \textbf{0.712} & 0.669 & 0.617 & \textbf{0.712} & 0.613 & 0.647 & 0.652 & 0.613 & 0.668 & 0.550 & 0.635 \\
    & max $E_{\phi}~\uparrow$ & \textbf{0.867} & 0.822 & 0.611 & 0.850 & 0.814 & 0.826 & 0.846 & 0.845 & 0.804 & 0.743 & 0.830  \\
    & mean $E_{\phi}~\uparrow$ & 0.749 & 0.663 & 0.555 & \textbf{0.815} & 0.659 & 0.662 & 0.737 & 0.603 & 0.773 & 0.434 & 0.704    \\
    & $\mathcal{M}~\downarrow$ & \textbf{0.027} & 0.034 & 0.032 & 0.034 & 0.068 & 0.033 & 0.046 & 0.032 & 0.039 & 0.033 & 0.030 \\ 
  \hline
  \multirow{4}{*}{Sp./Director} 
    & $S_{\alpha}~\uparrow$ & 0.679 & 0.753 & 0.701 & 0.677 & 0.756 & \textbf{0.772} & 0.726 & 0.716 & 0.755 & 0.731 & 0.672   \\
    & max $E_{\phi}~\uparrow$ & 0.852 & 0.880 & 0.891 & 0.832 & 0.902 & 0.900 & 0.894 & 0.899 & 0.883 & \textbf{0.918} & 0.844   \\
    & mean $E_{\phi}~\uparrow$ & 0,735 & 0.729 & \textbf{0.852} & 0.773 & 0.849 & 0.810 & 0.681 & 0.744 & 0.774 & 0.718 & 0.768   \\
    & $\mathcal{M}~\downarrow$ & 0.031 & 0.038 & 0.032 & 0.037 & 0.029 & \textbf{0.028} & 0.034 & 0.029 & 0.037 & 0.031 & 0.030 \\ 
   \hline
  \multirow{4}{*}{Sp./ChineseAd} 
    & $S_{\alpha}~\uparrow$ & 0.601 & \textbf{0.645} & 0.551 & 0.477 & 0.631 & 0.630 & 0.605 & 0.553 & 0.542 & 0.569 & 0.595  \\
    & max $E_{\phi}~\uparrow$ & 0.895 & 0.883 & 0.908 & 0.695 & 0.906 & 0.908 & \textbf{0.913} & 0.910 & 0.899 & 0.910 & 0.850  \\
    & mean $E_{\phi}~\uparrow$ & 0.483 & 0.524 & 0.527 & 0.391 & 0.576 & 0.635 & 0.544 & \textbf{0.641} & 0.632 & 0.470 & 0.502   \\
    & $\mathcal{M}~\downarrow$ & 0.009 & 0.009 & 0.042 & 0.069 & 0.027 & 0.011 & 0.012 & 0.015 & 0.028 & 0.010 & \textbf{0.007}   \\ 
    \hline
  \multirow{4}{*}{Sp./Exhibition} 
    & $S_{\alpha}~\uparrow$ & 0.487 & 0.469 & 0.480 & 0.469 & 0.428 & 0.492 & 0.486 & 0.487 & 0.473 & \textbf{0.510} & 0.475 \\
    & max $E_{\phi}~\uparrow$ & 0.614 & 0.658 & 0.605 & 0.597 & 0.689 & 0.770 & 0.735 & \textbf{0.811} & 0.576 & 0.773 & 0.560 \\
    & mean $E_{\phi}~\uparrow$ & 0.486 & 0.365 & 0.460 & 0.350 & 0.270 & 0.508 & 0.459 & \textbf{0.514} & 0.349 & 0.512 & 0.329  \\
    & $\mathcal{M}~\downarrow$ & 0.013 & 0.061 & 0.009 & 0.042 & 0.139 & 0.011 & 0.013 & \textbf{0.008} & 0.040 & 0.014 & 0.024 \\ 
    \hline
  \multirow{4}{*}{Sp./PianoConvo} 
    & $S_{\alpha}~\uparrow$ & 0.577 & 0.652 & 0.579 & 0.607 & 0.639 & 0.636 & 0.586 & 0.603 & \textbf{0.718} & 0.509 & 0.632   \\
    & max $E_{\phi}~\uparrow$ & 0.871 & 0.851 & 0.847 & 0.875 & 0.880 & 0.858 & 0.885 & \textbf{0.888} & 0.880 & 0.882 & 0.863  \\
    & mean $E_{\phi}~\uparrow$ & 0.774 & 0.673 & 0.745 & 0.833 & \textbf{0.847} & 0.807 & 0.693 & 0.706 & 0.803 & 0.420 & 0.804  \\
    & $\mathcal{M}~\downarrow$ & 0.037 & 0.035 & 0.035 &  0.038 & 0.033 & 0.036 & 0.036 & 0.035 & \textbf{0.028} & 0.033 & 0.033  \\ 
    \hline
  \multirow{4}{*}{Sp./FilmingSite} 
    & $S_{\alpha}~\uparrow$ & 0.578 & 0.633 & 0.603 & 0.610 & 0.637 & \textbf{0.645} & 0.636 & 0.578 & 0.640 & 0.633 & 0.522  \\
    & max $E_{\phi}~\uparrow$ & 0.727 & 0.762 & 0.681 & 0.766 & 0.799 & 0.766 & 0.787 & 0.708 & 0.738 & \textbf{0.805} & 0.793 \\
    & mean $E_{\phi}~\uparrow$ & 0.562 & 0.627 & 0.626 & 0.636 & 0.707 & 0.654 & 0.613 & 0.540 & 0.628 & 0.652 & \textbf{0.727} \\
    & $\mathcal{M}~\downarrow$ & 0.013 & 0.023 & 0.023 & 0.030 & 0.017 & 0.014 & 0.020 & \textbf{0.012} & 0.013 & 0.017 & 0.016   \\ 
    \hline
  \multirow{4}{*}{Sp./Brothers} 
    & $S_{\alpha}~\uparrow$ & 0.673 & 0.686 & 0.638 & 0.655 & 0.652 & \textbf{0.697} & 0.685 & 0.662 & 0.664 & 0.666 & 0.623 \\
    & max $E_{\phi}~\uparrow$ & 0.778 & 0.806 & 0.747 & 0.772 & 0.746 & 0.816 & 0.792 & 0.784 & 0.813 & \textbf{0.820} & 0.729 \\
    & mean $E_{\phi}~\uparrow$ & 0.688 & 0.677 & 0.713 & 0.705 & 0.706 & 0.715 & 0.629 & 0.661 & \textbf{0.728} & 0.650 & 0.681  \\
    & $\mathcal{M}~\downarrow$ & 0.018 & 0.024 & 0.023 & 0.024 & 0.025 & 0.017 & 0.022 & \textbf{0.015} & 0.019 & 0.016 & 0.016     \\ 
    \hline
  \multirow{4}{*}{Sp./Rap} 
    & $S_{\alpha}~\uparrow$ & 0.498 & 0.477 & 0.521 & 0.343 & 0.507 & 0.525 & 0.463 & 0.482 & 0.506 & 0.495 & \textbf{0.532}  \\
    & max $E_{\phi}~\uparrow$ & 0.830 & 0.816 & 0.831 & 0.471 & 0.814 & 0.824 & 0.761 & 0.828 & \textbf{0.858} & 0.832 & 0.818 \\
    & mean $E_{\phi}~\uparrow$ & 0.530 & 0.387 & 0.548 & 0.260 & 0.484 & 0.678 & 0.400 & 0.513 & 0.590 & 0.566 & \textbf{0.733} \\
    & $\mathcal{M}~\downarrow$ & \textbf{0.006} & 0.087 & 0.021 & 0.371 & 0.025 & 0.012 & 0.095 & 0.007 & 0.020 & 0.009 & 0.009 \\ 
    \hline
  \multirow{4}{*}{Sp./Spanish} 
    & $S_{\alpha}~\uparrow$ & 0.606 & 0.765 & 0.746 & 0.679 & \textbf{0.793} & 0.713 & 0.701 & 0.724 & 0.700 & 0.543 & 0.602  \\
    & max $E_{\phi}~\uparrow$ & 0.838 & 0.870 & 0.851 & 0.819 & 0.873 & 0.854 & 0.865 & \textbf{0.877} & 0.862 & 0.839 & 0.728 \\
    & mean $E_{\phi}~\uparrow$ & 0.651 & 0.807 & 0.835 & 0.727 & \textbf{0.865} & 0.797 & 0.822 & 0.784 & 0.800 & 0.486 & 0.503 \\
    & $\mathcal{M}~\downarrow$ & 0.038 & 0.030 & 0.032 & 0.035 & \textbf{0.025} & 0.036 & 0.040 & 0.032 & 0.037 & 0.042 & 0.035  \\ 
    \hline
  \multirow{4}{*}{Sp./Questions} 
    & $S_{\alpha}~\uparrow$ & 0.505 & 0.640 & \textbf{0.740} & 0.563 & 0.605 & 0.691 & 0.671 & 0.576 & 0.676 & 0.595 & 0.549 \\
    & max $E_{\phi}~\uparrow$ & 0.925 & 0.921 & 0.901 & 0.926 & 0.920 & 0.915 & 0.922 & \textbf{0.935}  & 0.907 & 0.909 & 0.757 \\
    & mean $E_{\phi}~\uparrow$ & 0.763 & 0.609 & \textbf{0.870} & 0.576 & 0.855 & 0.700 & 0.574 & 0.569 & 0.667 & 0.540 & 0.703 \\
    & $\mathcal{M}~\downarrow$ & 0.009 & 0.011 & \textbf{0.006} & 0.007 & 0.010 & 0.010 & 0.009 & 0.009 & 0.014 & 0.012 & 0.013 \\ 
    \hline
  \multirow{4}{*}{Sp./PianoMono} 
    & $S_{\alpha}~\uparrow$ & 0.598 & 0.555 & 0.573 & 0.572 & 0.629 & 0.522 & \textbf{0.637} & 0.506 & 0.611 & 0.502 & 0.502 \\
    & max $E_{\phi}~\uparrow$ & 0.702 & 0.855 & 0.766 & 0.796 & \textbf{0.861} & 0.842 & 0.755 & 0.859 & 0.831 & 0.796 & 0.715 \\
    & mean $E_{\phi}~\uparrow$ & 0.682 & 0.736 & 0.688 & 0.739 & 0.746 & \textbf{0.758} & 0.696 & 0.500 & 0.736 & 0.397 & 0.633 \\
    & $\mathcal{M}~\downarrow$ & 0.057 & 0.054 & 0.044 & 0.056 & 0.054 & 0.048 & 0.060 & 0.039 & 0.047 & \textbf{0.037} & 0.043  \\ 
    \hline
  \multirow{4}{*}{Sp./Snowfield} 
    & $S_{\alpha}~\uparrow$ & 0.729 & 0.811 & 0.778 & 0.800 & 0.819 & 0.779 & \textbf{0.823} & 0.601 & 0.794 & 0.580 & 0.578 \\
    & max $E_{\phi}~\uparrow$ & 0.739 & 0.816 & 0.741 & 0.784 & 0.819 & 0.763 & 0.836 & \textbf{0.864} & 0.779 & 0.812 & 0.775 \\
    & mean $E_{\phi}~\uparrow$ & 0.677 & 0.778 & 0.720 & 0.764 & \textbf{0.792} & 0.716 & 0.788 & 0.485 & 0.753 & 0.514 & 0.618 \\
    & $\mathcal{M}~\downarrow$ & 0.032 & 0.029 & 0.031 & 0.028 & \textbf{0.026} & 0.029 & 0.027 & 0.033 & 0.030 & 0.035 & 0.040  \\ 
    \hline
  \multirow{4}{*}{Sp./Melodrama} 
    & $S_{\alpha}~\uparrow$ & 0.609 & \textbf{0.685} & 0.655 & 0.673 & 0.667 & 0.664 & 0.617 & 0.467 & 0.608 & 0.604 & 0.568  \\
    & max $E_{\phi}~\uparrow$ & 0.782 & 0.835 & 0.837 & 0.811 & 0.835 & \textbf{0.841} & 0.816 & 0.788 & 0.816 & 0.831 & 0.794 \\
    & mean $E_{\phi}~\uparrow$ & 0.699 & 0.744 & 0.732 & 0.773 & \textbf{0.784} & 0.717 & 0.710 & 0.296 & 0.730 & 0.521 & 0.770 \\
    & $\mathcal{M}~\downarrow$ & 0.108 & 0.084 & \textbf{0.068} & 0.083 & 0.079 & 0.095 & 0.100 & 0.076 & 0.100 & 0.079 & 0.098  \\ 
    \hline
  \multirow{4}{*}{Sp./Gymnasium} 
    & $S_{\alpha}~\uparrow$ & \textbf{0.551} & 0.514 & 0.492 & 0.501 & 0.501 & 0.507 & 0.537 & 0.520 & 0.520 & 0.503 & 0.505  \\
    & max $E_{\phi}~\uparrow$ & 0.806 & 0.683 & 0.830 & 0.700 & 0.813 & 0.686 & 0.754 & \textbf{0.863} & 0.760 & 0.798 & 0.752  \\
    & mean $E_{\phi}~\uparrow$ & 0.584 & 0.545 & 0.461 & 0.593 & 0.469 & 0.512 & 0.487 & 0.584 & 0.518 & 0.468 & \textbf{0.642} \\
    & $\mathcal{M}~\downarrow$ & \textbf{0.007} & 0.013 & 0.021 & 0.011 & 0.020 & 0.016 & 0.020 & \textbf{0.007} & 0.017 & 0.027 & 0.010  \\
    \hline
  \multirow{4}{*}{Mu./Studio} 
    & $S_{\alpha}~\uparrow$ & 0.741 & 0.770 & 0.753 & \textbf{0.788} & 0.758 & 0.739 & 0.724 & 0.637 & 0.778 & 0.756 & 0.760  \\
    & max $E_{\phi}~\uparrow$ & 0.878 & 0.889 & 0.898 & \textbf{0.904} & 0.892 & 0.899 & 0.891 & \textbf{0.904} & 0.893 & 0.901 & 0.895 \\
    & mean $E_{\phi}~\uparrow$ & 0.745 & 0.731 & 0.832 & 0.826 & 0.847 & 0.756 & 0.601 & 0.629 & 0.800 & 0.729 & \textbf{0.859} \\
    & $\mathcal{M}~\downarrow$ & 0.008 & 0.009 & 0.010 & \textbf{0.006} & 0.009 & 0.009 & 0.010 & 0.008 & 0.009 & 0.008 & 0.007  \\ 
  
  \bottomrule
  \end{tabular}
  }
\end{table*}

\begin{table*}[t!]
  \centering
  \renewcommand{\arraystretch}{1.0}
  \setlength\tabcolsep{1.3pt}
  \caption{
   Sequence performance comparison of 7/3/1 SOTA I-SOD/V-SOD/PI-SOD methods. Mu. = Music. Mi. = Miscellanea.
   }\label{tab:SeqQua_2}
  \resizebox{0.99\textwidth}{!}{
  \begin{tabular}{l|r||ccccccc|ccc|c}
   \toprule
   \multirow{2}{*}{Super-class/Sequence}&\multirow{2}{*}{Metrics} & \multicolumn{7}{c|}{I-SOD} & \multicolumn{3}{c|}{V-SOD} & \multicolumn{1}{c}{PI-SOD}
  \\
  \cline{3-13}
   && CPD \cite{CPD}~ & SCRN \cite{SCRN}~ & F3Net \cite{F3Net}~ & MINet \cite{MINet}~ & LDF \cite{CVPR2020LDF}~ & CSF \cite{SOD100K}~ & GateNet \cite{GateNet}~ & COSNet \cite{COSNet}~ & RCRNet \cite{RCRNet}~ & PCSA \cite{gu2020PCSA}~ & FANet \cite{huang2020fanet}
   \\
  \hline
  \multirow{4}{*}{Mu./Church} 
    & $S_{\alpha}~\uparrow$ & 0.527 & 0.589 & 0.621 & 0.566 & 0.518 & 0.624 & 0.651 & 0.562 & 0.676 & 0.623 & \textbf{0.679} \\
    & max $E_{\phi}~\uparrow$ & 0.868 & 0.917 & 0.933 & 0.747 & 0.932 & 0.903 & \textbf{0.950} & 0.887 & 0.900 & 0.942 & 0.866 \\
    & mean $E_{\phi}~\uparrow$ & 0.451 & 0.575 & 0.731 & 0.576 & 0.715 & 0.601 & 0.657 & 0.487 & 0.635 & 0.577 & \textbf{0.774}  \\
    & $\mathcal{M}~\downarrow$ & 0.007 & 0.012 & 0.012 & 0.021 & 0.018 & 0.011 & 0.008 & \textbf{0.006} & 0.008 & 0.018 & 0.007   \\ 
    \hline
  \multirow{4}{*}{Mu./Duet} 
    & $S_{\alpha}~\uparrow$ & 0.662 & 0.704 & 0.698 & 0.653 & \textbf{0.751} & 0.648 & 0.730 & 0.553 & 0.731 & 0.538 & 0.643  \\
    & max $E_{\phi}~\uparrow$ & 0.879 & 0.891 & 0.892 & 0.876 & 0.898 & \textbf{0.903} & 0.883 & 0.901 & 0.889 & 0.873 & 0.876 \\
    & mean $E_{\phi}~\uparrow$ & 0.810 & 0.705 & 0.792 & \textbf{0.821} & 0.808 & 0.693 & 0.735 & 0.542 & 0.776 & 0.509 & 0.765  \\
    & $\mathcal{M}~\downarrow$ & 0.041 & 0.058 & 0.044 & 0.039 & 0.033 & 0.033 & 0.036 & 0.036 & 0.033 & 0.036 & \textbf{0.031} \\ 
  \hline
  \multirow{4}{*}{Mu./Blues} 
    & $S_{\alpha}~\uparrow$ & 0.580 & 0.742 & \textbf{0.776} & 0.722 & 0.771 & 0.734 & 0.740 & 0.595 & 0.765 & 0.743 & 0.600  \\
    & max $E_{\phi}~\uparrow$ & 0.879 & 0.889 & 0.890 & 0.802 & 0.871 & 0.844 & 0.893 & 0.884 & 0.894 & \textbf{0.904} & 0.852  \\
    & mean $E_{\phi}~\uparrow$ & 0.598 & 0.688 & 0.830 & 0.698 & \textbf{0.789} & 0.766 & 0.640 & 0.473 & 0.834 & 0.715 & 0.612 \\
    & $\mathcal{M}~\downarrow$ & 0.016 & 0.015 & \textbf{0.013} & 0.027 & 0.015 & 0.015 & 0.015 & 0.015 & \textbf{0.013} & 0.017 & 0.014  \\ 
  \hline
  \multirow{4}{*}{Mu./Violins} 
    & $S_{\alpha}~\uparrow$ & 0.589 & 0.668 & 0.537 & \textbf{0.692} & 0.661 & 0.631 & 0.656 & 0.578 & 0.669 & 0.671 & 0.604  \\
    & max $E_{\phi}~\uparrow$ & 0.852 & 0.877 & 0.578 & 0.861 & \textbf{0.883} & 0.845 & 0.872 & 0.851 & 0.868 & 0.856 & 0.790 \\
    & mean $E_{\phi}~\uparrow$ & 0.649 & 0.655 & 0.477 & \textbf{0.775} & 0.722 & 0.724 & 0.569 & 0.597 & 0.724 & 0.625 & 0.749 \\
    & $\mathcal{M}~\downarrow$ & 0.017 & 0.020 & \textbf{0.015} & 0.016 & 0.022 & 0.019 & 0.017 & \textbf{0.015} & 0.021 & 0.020 & 0.017  \\ 
   \hline
  \multirow{4}{*}{Mu./SingingDancing} 
    & $S_{\alpha}~\uparrow$ & 0.506 & \textbf{0.601} & 0.582 & 0.560 & 0.561 & 0.594 & 0.568 & 0.521 & 0.569 & 0.558 & 0.557  \\
    & max $E_{\phi}~\uparrow$ & 0.804 & \textbf{0.820} & 0.815 & \textbf{0.820} & 0.673 & 0.782 & 0.813 & 0.791 & 0.810 & 0.812 & 0.759 \\
    & mean $E_{\phi}~\uparrow$ & 0.500 & 0.587 & \textbf{0.758} & 0.565 & 0.618 & 0.589 & 0.547 & 0.452 & 0.637 & 0.608 & 0.705 \\
    & $\mathcal{M}~\downarrow$ & 0.026 & 0.034 & 0.037 & 0.025 & 0.042 & 0.026 & 0.026 & \textbf{0.023} & 0.030 & 0.034 & 0.033  \\ 
    \hline
  \multirow{4}{*}{Mi./Dog} 
    & $S_{\alpha}~\uparrow$ & 0.497 & 0.516 & \textbf{0.571} & 0.560 & 0.569 & 0.557 & 0.562 & 0.523 & 0.562 & 0.539 & 0.520  \\
    & max $E_{\phi}~\uparrow$ & 0.548 & 0.685 & 0.535 & 0.551 & 0.593 & 0.572 & 0.589 & 0.671 & 0.612 & \textbf{0.693} & 0.345 \\
    & mean $E_{\phi}~\uparrow$ & 0.460 & 0.457 & 0.493 & 0.511 & 0.424 & 0.532 & 0.515 & 0.494 & \textbf{0.548} & 0.544 & 0.325 \\
    & $\mathcal{M}~\downarrow$ & 0.013 & 0.015 & 0.014 & 0.007 & 0.020 & 0.004 & 0.009 & 0.005 & 0.004 & 0.005 & \textbf{0.003} \\ 
    \hline
  \multirow{4}{*}{Mi./RacingCar} 
    & $S_{\alpha}~\uparrow$ & 0.770 & 0.769 & 0.763 & 0.770 & 0.772 & 0.771 & \textbf{0.791} & 0.760 & 0.772 & 0.755 & 0.762  \\
    & max $E_{\phi}~\uparrow$ & 0.428 & 0.438 & 0.365 & 0.439 & \textbf{0.459} & 0.458 & 0.448 & 0.449 & 0.453 & 0.426 & 0.285 \\
    & mean $E_{\phi}~\uparrow$ & 0.315 & 0.338 & 0.283 & 0.349 & 0.332 & 0.287 & \textbf{0.370} & 0.276 & 0.293 & 0.261 & 0.260 \\
    & $\mathcal{M}~\downarrow$ & 0.089 & 0.115 & 0.087 & 0.109 & 0.102 & 0.085 & 0.107 & 0.083 & 0.087 & 0.085 & \textbf{0.081}  \\ 
    \hline
  \multirow{4}{*}{Mi./Train} 
    & $S_{\alpha}~\uparrow$ & 0.604 & 0.616 & 0.614 & 0.607 & 0.629 & 0.594 & \textbf{0.663} & 0.501 & 0.524 & 0.515 & 0.489 \\
    & max $E_{\phi}~\uparrow$ & 0.618 & 0.700 & 0.531 & 0.676 & 0.589 & 0.665 & \textbf{0.780} & 0.556 & 0.671 & 0.526 & 0.432 \\
    & mean $E_{\phi}~\uparrow$ & 0.581 & 0.553 & 0.486 & 0.493 & 0.554 & 0.558 & \textbf{0.634} & 0.351 & 0.462 & 0.416 & 0.386 \\
    & $\mathcal{M}~\downarrow$ & 0.024 & 0.030 & 0.020 & 0.041 & \textbf{0.012} & 0.013 & 0.022 & 0.016 & 0.016 & 0.028 & 0.016  \\ 
    \hline
  \multirow{4}{*}{Mi./Football} 
    & $S_{\alpha}~\uparrow$ & 0.653 & 0.696 & 0.618 & 0.656 & 0.668 & 0.658 & 0.676 & 0.648 & \textbf{0.710} & 0.635 & 0.556  \\
    & max $E_{\phi}~\uparrow$ & 0.833 & 0.856 & 0.790 & 0.830 & 0.835 & 0.846 & 0.820 & 0.811 & \textbf{0.866} & 0.810 & 0.742 \\
    & mean $E_{\phi}~\uparrow$ & 0.634 & 0.676 & 0.755 & 0.633 & \textbf{0.770} & 0.721 & 0.663 & 0.649 & 0.732 & 0.630 & 0.477 \\
    & $\mathcal{M}~\downarrow$ & 0.004 & 0.004 & 0.004 & 0.003 & 0.004 & 0.003 & 0.004 & 0.003 & \textbf{0.002} & 0.003 & \textbf{0.002} \\ 
    \hline
  \multirow{4}{*}{Mi./ParkingLot} 
    & $S_{\alpha}~\uparrow$ & 0.635 & 0.627 & 0.624 & 0.564 & \textbf{0.640} & 0.562 & 0.625 & 0.548 & 0.624 & 0.501 & 0.627  \\
    & max $E_{\phi}~\uparrow$ & \textbf{0.666} & 0.645 & 0.649 & 0.614 & 0.650 & 0.646 & 0.663 & 0.659 & 0.661 & 0.622 & 0.665 \\
    & mean $E_{\phi}~\uparrow$ & \textbf{0.641} & 0.551 & 0.600 & 0.597 & 0.625 & 0.602 & 0.610 & 0.482 & 0.612 & 0.501 & 0.593 \\
    & $\mathcal{M}~\downarrow$ & 0.028 & 0.041 & 0.048 & 0.059 & 0.041 & 0.035 & 0.045 & 0.027 & 0.038 & 0.029 & \textbf{0.026}   \\ 
    \hline
  \multirow{4}{*}{Mi./Skiing} 
    & $S_{\alpha}~\uparrow$ & 0.697 & 0.728 & 0.689 & 0.727 & 0.632 & \textbf{0.757} & 0.695 & 0.624 & 0.745 & 0.641 & 0.590  \\
    & max $E_{\phi}~\uparrow$ & 0.784 & 0.764 & 0.782 & 0.730 & \textbf{0.829} & 0.781 & 0.761 & 0.814 & 0.806 & 0.781 & 0.744 \\
    & mean $E_{\phi}~\uparrow$ & 0.705 & 0.645 & 0.669 & 0.661 & 0.517 & 0.675 & 0.605 & 0.573 & \textbf{0.716} & 0.613 & 0.500 \\
    & $\mathcal{M}~\downarrow$ & 0.015 & 0.024 & 0.027 & 0.025 & 0.044 & 0.015 & 0.030 & 0.014 & 0.016 & 0.016 & \textbf{0.012}  \\ 
    
  \bottomrule
  \end{tabular}
  }
\end{table*}

\begin{figure*}[t!]
	\centering
	\begin{overpic}[width=\textwidth]{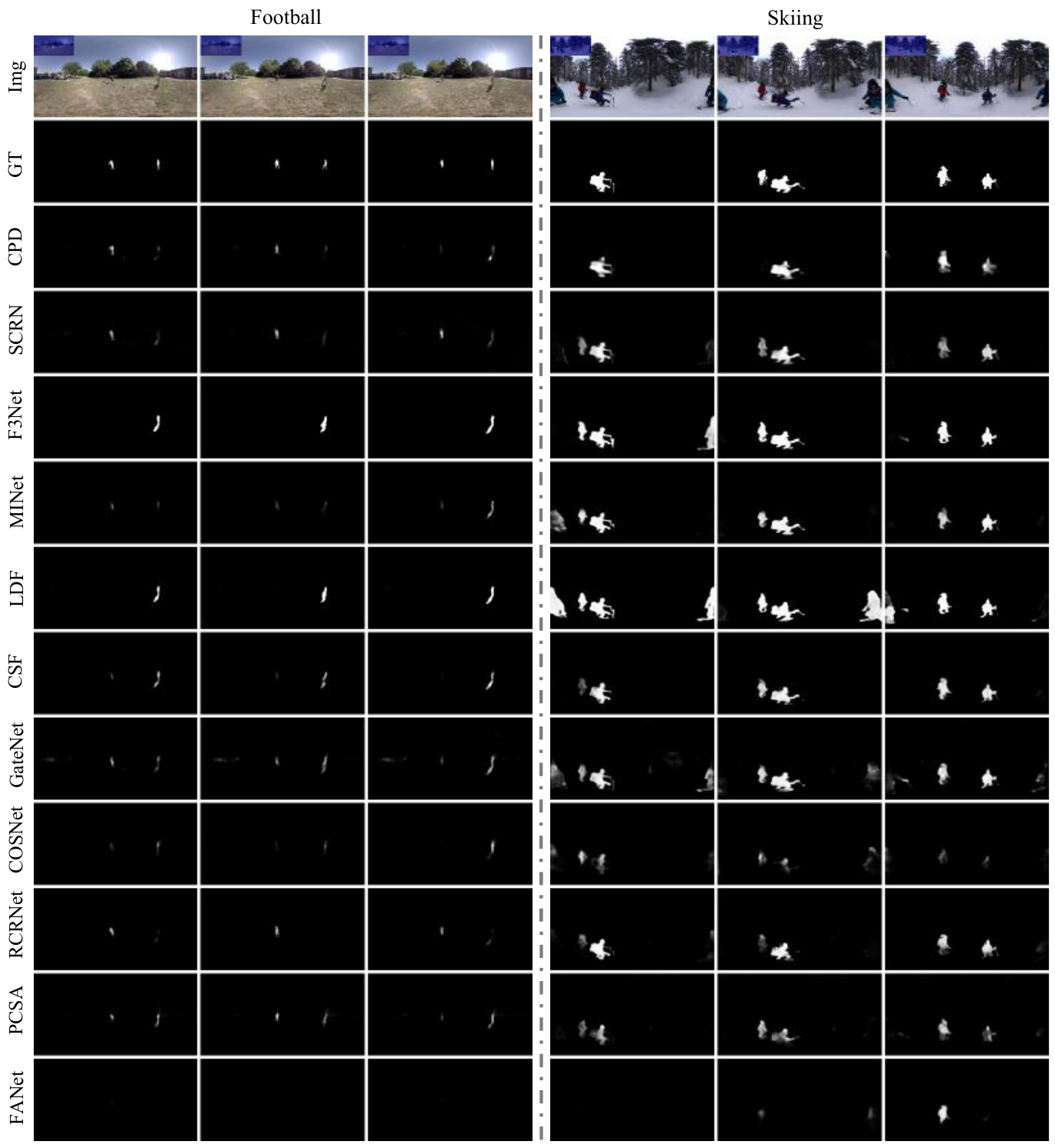}
    \end{overpic}
	\caption{Visual results of all baselines on the \ourdataset-test0 (Miscellanea). Img = image. GT = ground truth.}
    \label{fig:visual_te0}
\end{figure*}

\begin{figure*}[t!]
	\centering
	\begin{overpic}[width=\textwidth]{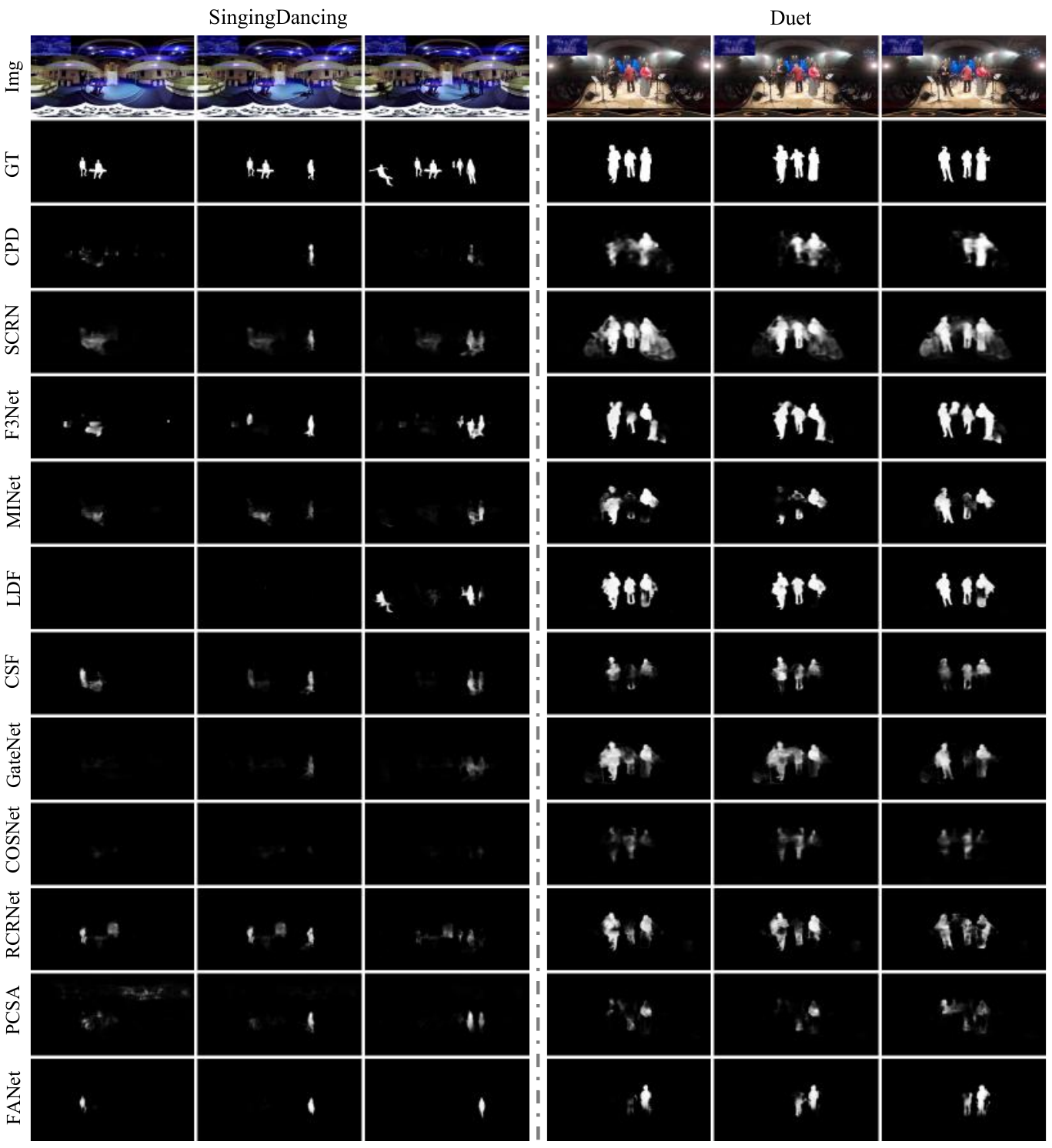}
    \end{overpic}
	\caption{Visual results of all baselines on the \ourdataset-test1 (Music). Img = image. GT = ground truth.}
    \label{fig:visual_te1}
\end{figure*}

\begin{figure*}[t!]
	\centering
	\begin{overpic}[width=\textwidth]{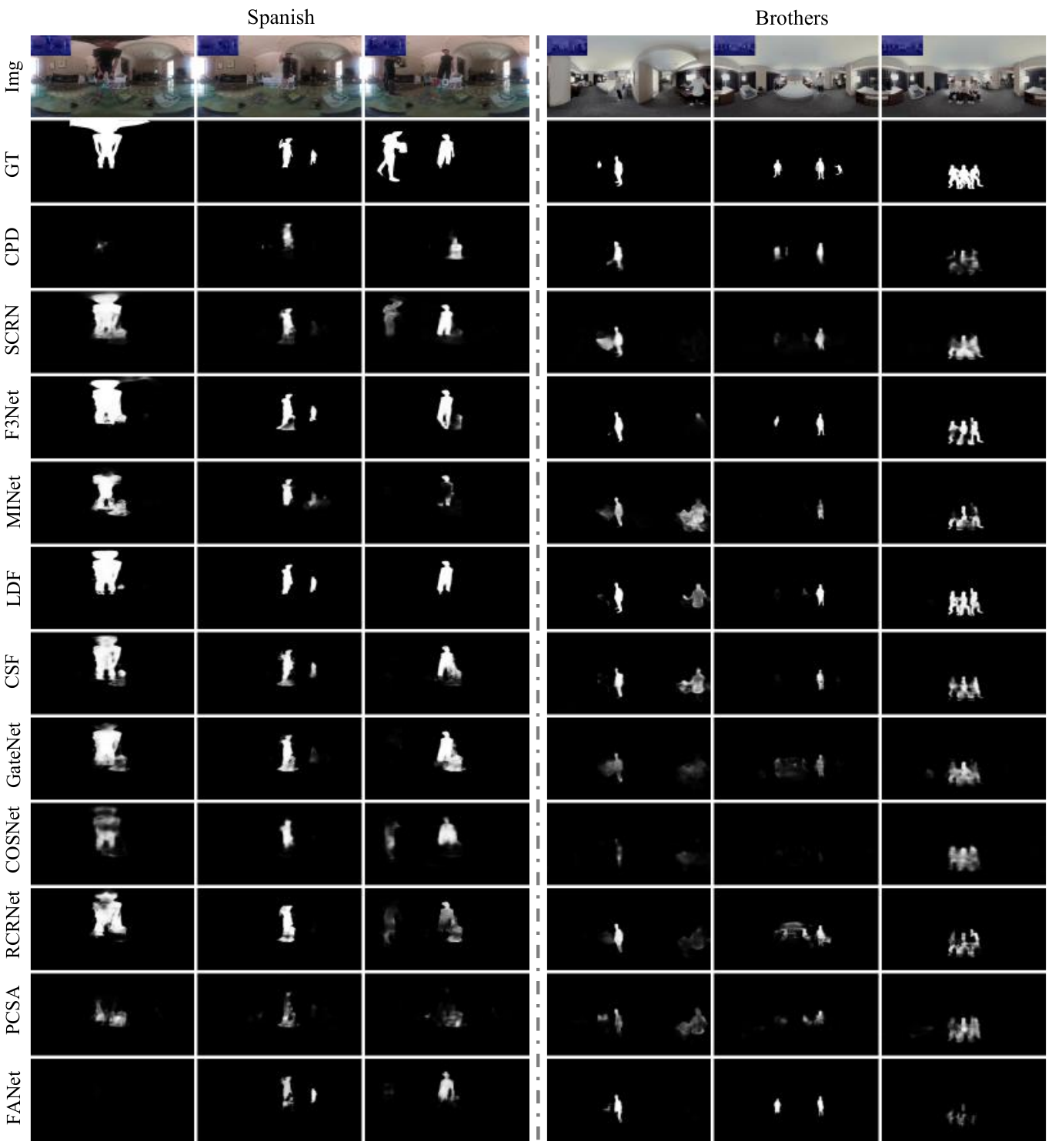}
    \end{overpic}
	\caption{Visual results of all baselines on the \ourdataset-test2 (Speaking). Img = image. GT = ground truth.}
    \label{fig:visual_te2}
\end{figure*}

% \begin{figure*}[t!]
% 	\centering
% 	\begin{overpic}[width=.96\textwidth]{fig_show_part2.pdf}
%     \end{overpic}
% 	\caption{Sample key frames from \ourdataset, with fixations and instance-level ground truth overlaid.}\label{fig:show_part2}
% \end{figure*}

\section*{Appendix}
\noindent
\textbf{Per Video Performance \& Visual Results.}
The per-video quantitative results are shown in \tabref{tab:SeqQua_1} and \tabref{tab:SeqQua_2}. Please refer to \figref{fig:visual_te0}. \figref{fig:visual_te1} and \figref{fig:visual_te2} for visual results. 

\bibliographystyle{CVM}

{\normalsize  \bibliography{CVM}}

\clearpage
\Author{ZhangYi}{Yi Zhang}
{received his bachelor degree in 2016 and master degree in 2019 respectively from Southeast University both in biomedical engineering. He is pursuing his PhD degree at INSA Rennes, France. His current research interests include omnidirectional vision, salient object detection, camouflaged object detection and deep learning.}

\end{document}